\documentclass[10pt]{article}
\pdfoutput=1
\usepackage{graphicx,graphics}
\usepackage[square,numbers]{natbib} 
\DeclareGraphicsExtensions{.pdf,.png,.jpg}
\usepackage{xspace,xcolor}
\usepackage{fullpage}
\usepackage{amsmath, amsthm, amsfonts,mathrsfs}
\usepackage[linesnumbered, ruled, vlined, algo2e]{algorithm2e} % , 
\usepackage{times}
\usepackage{mathtools}
\usepackage{enumitem}
\usepackage{footnote}
\usepackage{float}
\usepackage{multirow}
\usepackage{nicefrac}
\usepackage{wrapfig}
\usepackage{framed}
\usepackage{url}
\usepackage[skip=0pt]{adjustbox}
\usepackage{caption}
\usepackage{wrapfig}
\usepackage{dblfloatfix} 
\usepackage{tikz}
\usetikzlibrary{arrows,chains,matrix,positioning,scopes}
\usepackage{cases}
\usepackage[overload]{empheq}
\DeclareMathAlphabet{\mathpzc}{OT1}{pzc}{m}{it}
\SetCommentSty{mycommfont}

\usepackage{hyperref}
\usepackage{natbib}

\usepackage{makecell}
\usepackage[american]{babel}
\usepackage{lscape}
\usepackage{subfig}

\newcommand{\ie}{i.e.\ }
\newcommand{\eg}{e.g.\ }

\newcommand{\TODO}[1]{{\color{red} TODO: #1}}

\newcommand{\eat}[1]{}
\usepackage{amssymb,latexsym}
\usepackage{amsmath}
\usepackage{amsfonts}
\usepackage{amsthm,amsopn}

\usepackage{bm,bbm} % bold symbol

\newcommand{\EI}{\textsf{EI}\xspace}
\newcommand{\BO}{BO\xspace}

\newcommand{\HB}{Hyperband\xspace}

\newcommand{\cifar}{CIFAR-10\xspace}

\newcommand{\e}{y}
\newcommand{\m}{k}

\newcommand{\mO}{\mat{\mathrm{O}}}
\newcommand{\mK}{\mat{\mathrm{K}}}
\newcommand{\mS}{\mathrm{S}}
\newcommand{\mQ}{\mathrm{Q}}

\newcommand{\vct}[1]{\boldsymbol{#1}} % vector
\newcommand{\mat}[1]{\boldsymbol{#1}} % matrix
\newcommand{\cst}[1]{\mathsf{#1}}  % constant

\newcommand{\field}[1]{\mathbb{#1}}
\newcommand{\R}{\field{R}} % real domain
\newcommand{\T}{^{\top}\!\!} % transpose
\newcommand\given[1][]{\:#1\vert\:}

\newcommand{\ProbOpr}[1]{\mathbb{#1}}
\newcommand{\expect}[2]{%
\ifthenelse{\equal{#2}{}}{\ProbOpr{E}_{#1}}
{\ifthenelse{\equal{#1}{}}{\ProbOpr{E}\left[#2\right]}{\ProbOpr{E}_{#1}\left[#2\right]}}} 
\newcommand{\var}[2]{%
\ifthenelse{\equal{#2}{}}{\ProbOpr{VAR}_{#1}}
{\ifthenelse{\equal{#1}{}}{\ProbOpr{VAR}\left[#2\right]}{\ProbOpr{VAR}_{#1}\left[#2\right]}}} % Expectation: syntax: V{1}{2} = V_1[2], V{}{2}=V[2], V{1}{} = V_1
\newcommand{\std}[2]{%
\ifthenelse{\equal{#2}{}}{\ProbOpr{\sigma}_{#1}}
{\ifthenelse{\equal{#1}{}}{\ProbOpr{\sigma}\left[#2\right]}{\ProbOpr{\sigma}_{#1}\left[#2\right]}}}

\DeclareMathOperator*{\argmin}{argmin}

\newcommand{\diag}{\operatornamewithlimits{diag}}

\newtheorem{example}{Example}

\newcommand{\vf}{\vct{f}}

\newcommand{\vt}{\vct{t}}
\newcommand{\vx}{{\vct{x}}}
\newcommand{\vy}{\vct{y}}

\newcommand{\ones}{\vct{1}}

\eat{

}

\begin{document}

\title{Hyper-parameter Tuning under a Budget Constraint}
\author{Zhiyun Lu, Chao-Kai Chiang, Fei Sha\\
Dept. of Computer Science,  U. of Southern California, Los Angeles, CA 90089\\
 \texttt{\{zhiyunlu.is.alive, chaokai\}@gmail.com, feisha@usc.edu}\\
}

\date{}
\maketitle

%!TEX root = main.tex

\begin{abstract}
We study a budgeted hyper-parameter tuning problem, where we optimize the tuning result under a hard resource constraint. 
We propose to solve it as a sequential decision making problem, such that we can use the partial training progress of configurations to dynamically allocate the remaining budget.
Our algorithm combines a Bayesian belief model which estimates the future performance of configurations, with an action-value function which balances exploration-exploitation tradeoff,
to optimize the final output. It automatically adapts the tuning behaviors to different constraints, which is useful in practice. Experiment results demonstrate superior performance over existing algorithms, including the-state-of-the-art one, on real-world tuning tasks across a range of different budgets.
\end{abstract}
% suspends and resumes the training of configurations
\eat{decide where to allocate the budget next.}

%!TEX root = main.tex
\section{Introduction}
Hyper-parameter tuning is of crucial importance to designing and deploying machine learning systems. Broadly, hyper-parameters include the architecture of the learning models, regularization parameters, optimization methods and their parameters, and other ``knobs'' to be tuned. 
It is challenging to explore the vast space of hyper-parameters efficiently to identify the optimal configuration.
Quite a few approaches have been proposed and investigated: random search,  Bayesian Optimization (BO) ~\citep{snoek2012practical,shahriari2016taking}, bandits-based Hyperband~\citep{jamieson2016non,li2016hyperband}, and meta-learning~\citep{chen2016learning,bello2017neural, franceschi2018bilevel}.

Many of those prior studies have focused on the aspect of reducing as much as possible the computation cost to obtain the optimal configuration. 
In this work, we look at a different but important perspective to hyper-parameter optimization -- under a fixed time/computation cost, how we can improve the performance as much as possible.
Concretely, we study the problem of hyper-parameter tuning under a budget constraint. 
The budget offers the practitioners a tradeoff: affordable time and resource balanced with models that are good -- \emph{best models that one can afford}. Often, this is a more realistic scenario in developing large-scale learning systems, and is especially applicable, for example when the practitioner searches for a best model under the pressure of a deadline. 

The budget constraint certainly complicates the hyper-parameter tuning strategy. While the strategy without the constraints is to explore and exploit in the hyper-parameter configuration space, a \emph{budget-aware} strategy needs to decide how much to explore and exploit with respect to the resource/time. 
As most learning algorithms are iterative in nature, a human operator would monitor the training progress of different configurations, 
and make judgement calls on their potential future performance,
based on what the tuning procedure has achieved so far, and how much resource remains. 
For example, as the deadline approaches, he/she might decide to exploit current best configuration to further establish its performance, instead of exploring a potentially better configuration as if he/she had unlimited time. 
\emph{Then how can we automate this process?}

We formalize this inquiry into a sequential decision making problem, and propose an algorithm to automatically achieve good resource utilization in the tuning. 
The algorithm uses a belief model to predict future performances of configurations. 
We design an action-value function, inspired by the idea of the Value of Information~\citep{dearden1998bayesian}, to select the configurations. 
The action-value function balances the tradeoff between exploration -- gather information to estimate the training curves, and exploitation -- achieve good performance under the budget constraint.

We empirically demonstrate the performance of the proposed algorithm on both synthetic and real-world datasets.  Our algorithm outperforms the state-of-the-art hyper-parameter tuning algorithms across a range of budgets. Besides, it exhibits budget adaptive tuning behaviors.

The rest of the paper is organized as follows.
In Section~\ref{sec:problem}, we formally define the problem  and introduce the sequential decision making formulatio. In Section~\ref{sec:approach}, we describe the proposed algorithm with analysis. Related work, experiments, and the conclusion can be found in Section~\ref{sec:related}, ~\ref{sec:exp}, and~\ref{sec:conclusion} respectively. % 

\eat{
\TODO{discuss pros and cons of two settings; discuss `budget' notion in existing work; discuss the significance of the new setting.}}

\eat{
Specifically, \TODO{outline as follows}
\begin{itemize}
\item we formulate a sequential decision making problem.
\item we explain the optimal planning structure in hyper-parameter tuning, and leverage this structure in our solution.
\item inspired by the idea of value of information in Bayesian RL, we design an action-value function which can efficiently explore to identify the best configuration.
\item our approach handles the challenge of planning and learning separately,
compared to other approximate dynamic programming solutions
\end{itemize}
We empirically validate our proposed solution on synthetic and real-world datasets.

}

\eat{
which is a joint function of the number of different configurations we try and how long we optimize the model under each configuration. 
A reasonable strategy would be that we quickly try out different configurations, and then allocate the resource to the promising configurations, so that we can deliver a well-trained model at the end of the budget.

It consists of two components:  a belief model which predicts future performances of configurations; and  an action-value function, which takes our predictions as input and decide which configuration we would spend the next budget on.
}

%!TEX root = main.tex

\section{Problem Statement}
\label{sec:problem}
In this section, we start by introducing notations, and formally define the budgeted hyper-parameter tuning problem. Finally we formulate the budgeted tuning task into a sequential decision making problem.
\bgroup
\everymath{\displaystyle}

\subsection{Preliminaries}
\eat{
Imagine that we need to conduct tuning under some computation constraint. We have some candidate configurations we would like to invest in. Note that when we calculate the total cost of tuning, not only the number of different configurations, but also how long we optimize the model under each configuration counts. 
At the end of tuning, we select and output a single model which achieves the best performance, among all models from all configurations we have tuned.
We formally define the problem as follows.
}
\paragraph{Configuration (arm)} Configuration denotes the hyper-parameter setting, \eg the architecture, the optimization method. We use $[K] = \{1,2,...,K\}$ to index the set of configurations. \eat{We simplify the problem to a finite set of configurations to focus on the main challenge of budgeted tuning. }The term configuration and arm are used interchangeably. (See Appendix for a full notation table.)

\paragraph{Model} Model refers to the (intermediate) training outcome, \eg the weights of neural nets, of a particular configuration. We evaluate the model on a heldout set periodically, for example every epoch. We consider loss or error rate as the evaluation metric.
Assume we have run configuration $k$ for $b$ epochs, and we keep track of  the loss of the best model as  $\nu^k_b \in [0, 1)$, the \emph{minimum} loss from $b$ epochs. Note that $\nu^k_b$ is a non-increasing function in $b$. We always use superscript to denote the configuration, and the subscript for the budget/epoch.
%\begin{align}
%\nu_{b}^k = \min_{ t \leq b} \e^{\m}(t), \label{eq:long_term}
%\end{align}

\paragraph{Budget} The budget defines a computation constraint imposed on the tuning process. In this paper, we consider a training epoch as the budget unit\footnote{In practice, we can generalize to other definitions of budget units.}, as this is an abstract notion of computation resource in most empirical studies of iterative learning algorithms. 
Given a total budget $B \in \mathbb{N}_+$, a strategy $\vct{b} = (b^1, \ldots, b^K) \in \mathbb{N}^K$ allocates the budget among $K$ configurations, \ie $k$
runs $b^k$ epochs respectively. $\vct{b}$ should satisfy that the total epochs from $K$ arms add up to $B$: $\vct{b}\T \bm{1}_K = B$.
We use epoch and budget interchangeably when there is no confusion. 

\paragraph{Constrained Optimization} The goal of the budgeted hyper-parameter tuning task is to obtain a well-optimized model under the constraint. 
Under the allocation strategy $\vct{b}$, arm $k$ returns a model with loss $
\nu^k_{b^k}$. \eat{at the end of the budget.} %= \min_{ t \leq b^{\m}} \e^{\m}(t)  
We search for the strategy,  which optimizes the loss of the \emph{best} model out of  $K$ configurations, $\ell_B = \min\{  \nu^1_{b^1}, \ldots, \nu^K_{b^K}\}$. 
Concretely, the constrained optimization problem is 
\begin{align}
% \ell_B = \min_{\m }  \nu^k_{b^k}, & \quad
\min_{\vct{b} }\ell_B,  \ \text{s.t. } \vct{b}\T \bm{1}_K = B. \label{eq:obj} 
%\m^*(B) = \argmin_{\m } \min_{ t \leq B}\e^{\m}(t). \label{eq:obj-sol}
\end{align}

\subsection{Optimal Solution with Perfect Information} \label{sec:structure}
Despite a combinatorial optimization, if we \emph{know} the training curves of all configurations (known $\nu^k_b \ \forall k$ and $b\leq B$), the solution to Eq.~\ref{eq:obj} has a simple structure: (one of) the optimal planning path must be a greedy one -- the budget is invested on a single configuration, due to the non-increasing nature of $\nu_{b^k}^k$ w.r.t. $b^k$. 
Hence the problem becomes to find the optimal arm $c$, which attains the smallest loss after $B$ epochs. And the optimal planning is simply to invest all $B$ units to $c$. 
{\begin{align}
&  \ell_B^* = \min_k \nu_B^k, & c(B) = \argmin_k \nu_B^k. \label{eq:opt_arm}
\end{align}
% \ b^c = B, b^k = 0, \forall k\neq c   
% b^k = B \cdot \mathbbm{1}[ k= c]
}Namely, $\nu_B^c = \ell_B^*$. Note $c$ is budget dependent. However in practice, computing  Eq.~\ref{eq:opt_arm} is infeasible, since we need to know \emph{all} $K$ values of  $\nu_B^k$, which already uses up $B\times K$ epochs, exceeding the budget $B$!
In other words, the challenge of the budgeted tuning lies in that we try to attain the minimum loss (of $B\times K$ epochs) $\ell_B^*$, with partial information of the curves from only $B$ observations/epochs.

%% sequential decision making
\subsection{Sequential Decision Making}\label{sec:formulation}
The challenge of \emph{unknown} $\nu_B^k$ comes from that the training of the iterative learning algorithm is innate sequential -- we cannot know the $B$-th epoch loss \eat{best}$\nu^k_{B}$ without actual running the first $B-1$ epochs using up the budget. 
On the other hand, this challenge can be attacked naturally in the framework of sequential decision making: the partial training curve is indicative of its future, and thus informs us of decisions early on without wasting the budget to finish the whole curve -- we all have had the experience of looking at a training curve and ``kill'' the running job as ``this training is going nowhere''! The key insight is that we want to use the observed information from epochs in the past, to decide which configuration to run or stop in the future.
\bgroup
\everymath{\displaystyle}

This motivates us to formulate the budgeted tuning as a finite-horizon sequential decision making problem. The training curves can be seen as an oblivious adversary environment~\citep{bubeck2012regret}: 
loss sequences of $K$  configurations are pre-generated before the tuning starts. \eat{(independent of the tuning algorithm)}
At the $n$-th step, the tuning algorithm selects the action/configuration $a_n \in [K]$ for the next budget/epoch, and the curve returns the corresponding observation/loss $z_n \in [0, 1)$ from configuration $a_n$.\eat{\footnote{We use $y^k(t)$ to denote the loss from the $t$-th epoch of arm $k$. Then $z_n = y^{a_n}(t) $ for some $t$, based on actions prior to $n$. 
$n$ is the global clock of the sequential decision making problem, while $t$ is the local epoch of arm $k$.  The sequence $\{z_n\}$ is a global re-arrangement of losses $y^k(t)$ from $K$ arms based on $\{a_n\}$.}}
Hence we get  a trajectory of configurations and losses, $\xi_n = (a_1, z_1, \ldots, a_n, z_n)$.
We define the policy to select actions as $\pi$: $a_n = \pi (\xi_{n-1}) = \pi(a_1, z_1, \ldots, a_{n-1}, z_{n-1})$, a function from the history to the next arm.  
This process is repeated for $B$ steps until the budget exhausts, and the final tuning output is 
\begin{align}
\ell^\pi_B = \min_{1\leq n\leq B} z_n. \label{eq:sobj}
\end{align}
Note $\ell^\pi_B$ is the same as the original objective $\ell_B$ in Eq.~\ref{eq:obj}.

In this way, we can solve the budgeted optimization problem (Eq.~\ref{eq:obj}) in a sequential manner, such that we  can leverage the information from partial training curves to make better decisions.
The sequential formulation enables us to stop or resume the training of a configuration at any time, which is economical in budget. Besides, the algorithm can exploit the information of budget $B$ to make better judgements and decisions. The framework can be extended to other settings, \eg multiple configurations in parallel, resource of heterogenous types (like cpu, gpu), and etc. While some of the extensions is straightforward, some requires more careful design, which is left as future work.
%The selection of the vector b can be done in a seq. way as described here.
%\TODO{What's the advantage of the sequential formulation in hyper-parameter tuning? }
\eat{To summarize, we have formulated a sequential decision making problem to solve the budgeted tuning task. } 

\paragraph{Regret}
%\paragraph{Regret}
The goal of the sequential problem is to optimize the policy $\pi$, such that the regret of our output against the optimal solution (Eq.~\ref{eq:opt_arm}) is minimized,
\begin{align}
\min_{\pi} \ell^\pi_B - \ell^*_B = \min_{\pi} \Big(\min_{1\leq n\leq B} z_n - \nu^c_B \Big) \label{eq:regret}.
\end{align} 
Note $\ell^*_B$ assumes knowledge of all curves, which is infeasible in practice. 

\paragraph{Challenges} 
First of all, to optimize under unknown curves leads to the well-known exploitation-exploration tradeoff, the same as in multi-armed bandits. On one hand, it is tempting to pick configurations that have achieved small losses (exploitation), but on the other hand, there is also an incentive to pick other
configurations to see whether they can admit even smaller losses (exploration).

However, Eq.~\ref{eq:regret} is different than the typical bandits, because the output $\ell^\pi_B$ is in the form of the \emph{minimal} loss, instead of the sum of losses. 
This leads to two key differences of the optimal policy: firstly the optimal policy/configuration depends on the horizon $B$.
Secondly at every step, the optimal policy for the remaining horizon depends on the history actions. 
Therefore, the key step is to \emph{re-plan} given what has been done so far every time.

\section{Approach}
\label{sec:approach}

In this section, we describe our budgeted tuning algorithm which attains the same performance as $\ell_B^*$ asymptotically (Eq.~\ref{eq:regret}).
At every step, we re-plan for the remaining horizon by leveraging the optimal planning structure (Sect.~\ref{sec:structure}), and employ the idea of Value of Information~\citep{dearden1998bayesian} to handle the exploration exploitation tradeoff.

Specifically, we design an action-value function (Sec.~\ref{sec:action}) to select the next action/configuration. 
The action-value function takes future predictions of the curves as input, which we use a Bayesian belief model to compute (Sec.~\ref{sec:belief}), and output a score measuring some expected future performance. 
We obtain a new observation from the selected arm, and update the belief model. 
We discuss how the action-value function balances the exploration-exploitation tradeoff in Sec.~\ref{sec:behavior}, and present the algorithm in Sec.~\ref{sec:alg}.
Lastly, in Sec.~\ref{sec:dp}, we discuss and compare with the Bayes optimal solution.

\eat{a different objective of the sequential decision making problem, which leads to }
\eat{%The key  is that the losses from the same configuration are correlated and we can estimate the future performance based on past observations.
% how we utilize the correlation of losses from the training curves to solve
}
\eat{model to predict training curves' future performances, described in Sec.~\ref{sec:belief}. }
\eat{
The key is that it efficiently gathers information to identify configurations with good future performances (exploration), as well as plans and adjusts the decisions according to the budget exhaustion (exploitation). 
The action-value function balances the exploration and exploitation tradeoff to simultaneously improve the belief model and plan optimally to achieve small loss.  
The proposed algorithm has the following intuitive behavior: it tries out different configurations at the beginning, and then \eat{gradually} zooms into a group of promising configurations with good \emph{long-term} performances, until it settles to the best configuration and invests the rest budget to improve its loss until convergence. 
}

%!TEX root = main.tex
\subsection{Action-value Function $\mQ$}
\label{sec:action}
\bgroup
\everymath{\displaystyle}
Consider we are at the $n$-th step, with remaining budget $r = B-n$.
Our goal is to find the policy which minimizes the tail sequence in Eq.~\ref{eq:sobj}: $\ell_r = \min_{n\leq s \leq B} {z_s}$. We use a belief state/model $\mS_{n-1}$ to estimate the unknown training curves. $\mS_{n-1}$ is a random process derived from the past trajectory $\xi_{n-1}$, which allows us to \eat{simulate future outcomes, as well as }simulate, and predict future outcomes.
We compute an action-value function  $\mQ_r[a]$ for each arm, and select the next action by minimizing it, 
\begin{align}
a_n = \pi(\xi_{n-1}) = \argmin_{a\in[K]} \mQ_r[a | \mS_{n-1}]. \label{eq:policy1}
\end{align}
\eat{where $r$ is the remaining budget.  }We drop the $\mS_{n-1}$ in $\mQ_r[a]$ to simplify the notation when it's clear. 
In what follows, we define $\mQ$ as a form of expected best loss we are likely to obtain of $\ell_r$, if we follow configuration $a$.
%$\mQ_r[a | \mS_{n-1}]$ measures the best loss we are likely to obtain from $a$.

%\paragraph{Planning}
First of all, recall the optimal planning of hyper-parameter tuning in Sect.~\ref{sec:structure}, there are $K$ candidates $\nu_r^k$ (the minimum loss from $k$ in $r$ epochs\footnote{With a slight abuse of notation, we use $\nu_r^k$ to denote its best performance in $r$ more epochs, starting from the current epoch.}) for the optimal $\ell^*_r$. 
This structure effectively reduces the combinatorial search space\footnote{There are exactly ${{r+K-1}\choose{K-1}} K!$ different  possible outcomes (ignore the sequence ordering).} 
of $\pi$ to a constant factor of $K$.
Thus we use the predictive value of $\nu_r^a$ to construct $\mQ_r[a]$.

%As a result we are uncertain which $k$ should be picked as the best configuration $c$. 

Since the belief $\mS_{n-1}$ is uncertain about the true curves, there is a distribution over values of $\nu_r^k$. We would like do exploration to improve the estimates of $\nu_r^k$s. Note that our focus is to correctly identify the best arm, instead of estimating all arms equally well. Inspired by the idea of Value of Information exploration in Bayesian reinforcement learning~\cite{dearden1998bayesian,dearden1999model}, we quantify the gains of exploration  in $\mQ$, by the amount of improvement on future loss.

Specifically, what can be an informative outcome that updates the agent's knowledge and leads to an improved future loss?
There are two scenarios where the outcomes contradict to our prior belief: 
(a) when the new sample surprises us by showing that an
action previously considered sub-optimal turns out to be the best arm,
and (b) when the new sample indicates a surprise that an action
that previously considered best is actually inferior to
other actions. 

Before we formally define $\mQ_r[a]$, we introduce the following notations.
Define $\mu^k_r = \expect{}{\nu_r^k  | \mS_{n-1}}$ as $k$'s expected best future loss. We can sort all arms based on their expected loss, and we call $\widehat{c} = \argmin_k \mu_r^k$ the predicted top arm, our current guess of $c$ (Eq.~\ref{eq:opt_arm}). Denote $\mu^{\text{1st}}_r = \mu^{\widehat{c}}_r$ as the top arm's expected long-term performance, and $\mu^{\text{2nd}}_r $ that of the runner up. 

Now consider in case (a) for a sub-optimal arm $a \neq \widehat{c}$, it will update  our estimate of $\ell_r$ when the sample $ \nu^a_{r} < \mu_{r}^{\text{1st}}$ outperforms the previously considered best expected loss. We
expect to gain $\mu_{r}^{\text{1st}} - \nu^a_{r}$ by taking $a$ instead of $\widehat{c}$. Define
\eat{
We reward this unexpected improvement, \eat{over $\mu_{r}^{\text{1st}}$ by taking the smaller among the two} and define}
\begin{align}
\mQ_r[a] = \ProbOpr{E}\left[\min \{ \nu^a_{r},  \mu_{r}^{\text{1st}}\} \right]
= \mu_{r}^{\text{1st}} - \ProbOpr{E}\left[\left (\mu_{r}^{\text{1st}} - \nu^a_{r} \right)^+ \right].
% = \mu_{r}^{\text{1st}} - \Pr( \nu^a_{r} < \mu_{r}^{\text{1st}}) \ProbOpr{E}\left[\mu_{r}^{\text{1st}} - \nu^a_{r} \given[\Big]   \nu^a_{r} < \mu_{r}^{\text{1st}}\right],
\label{eq:Qsub}
\end{align}
The second term on the r.h.s. computes the area when $\nu^a_{r}$ falls smaller than $\mu_{r}^{\text{1st}}$, which is called the value of perfect information (VoI) in decision theory~\citep{howard1966information}. It is a numerical
value that measures the reduction of uncertainty, thus can assess the value for exploring action $a$.
In our problem, intuitively it quantifies the average surprise of $a$ over all draws from the belief. Minimizing $\mQ_r[a]$ favors $a$ with large surprise/VoI.
\eat{assigns credit to possible improvement from $a$. }
 
Similarly in case (b) when we consider the predicted top arm $a = \widehat{c}$, there is a surprise that the sample $ \nu_r^{\widehat{c}} > \mu_{r}^{\text{2nd}}$ falls behind with other candidates. We define
\begin{align}
\mQ_r[\widehat{c}] = \ProbOpr{E}\left[\min \{ \nu^{\widehat{c}}_{r}, \mu_{r}^{\mathrm{2nd}} \}\right]
=  \mu_{r}^{\mathrm{1st}} -  \ProbOpr{E}\left[\left (
\nu^{\widehat{c}}_{r} -  \mu_{r}^{\mathrm{2nd}}
\right)^+ \right],
%=\mu_{r}^{\mathrm{1st}} - \Pr(\nu_r^{\widehat{c}} > \mu_{r}^{\text{2nd}}) \ProbOpr{E}\left[\nu^{\widehat{c}}_{r} -  \mu_{r}^{\mathrm{2nd}} \given[\Big]  \nu_r^{\widehat{c}} > \mu_{r}^{\text{2nd}}\right]
\label{eq:Qopt}
\end{align}
where the second term computes the area when $\nu_r^{\widehat{c}}$ unluckily falls right to $\mu_{r}^{\text{2nd}}$. Continuing with $\widehat{c}$ is favorable if it has high VoI, \ie gains us much knowledge from this surprise. 
See Fig.~\ref{fig:viz_acq} in the Appendix~\ref{ap:visual} for visualization of the action-value function.

Combining Eq.~\ref{eq:Qsub} and ~\ref{eq:Qopt}, the action-value function is
\begin{align}
\mQ_r[a] &  =  \ProbOpr{E}\left[\min \{ \nu^a_{r},  \min_{k \neq a } \mu^k_{r} \} \right]   \label{eq:acq}
% \\
%&  =
%    \begin{cases}
%   \ProbOpr{E}\left[\min \{ \nu^a_{r},  \mu_{r}^{\text{1st}}\}\right], & \text{ when } a \neq \widehat{c}, 
%     \\
%   \ProbOpr{E}\left[\min \{ \nu^{\widehat{c}}_{r},  \mu_{r}^{\mathrm{2nd}} \}\right],   &
%    \text{ when }  a = \widehat{c}. 
%    \end{cases}
%   
\end{align}
Finally, note that $\mQ$ depends on the remaining horizon $r$, through the index in $\nu^k_r$ and $\mu^k_r$, and the history  $\xi_{n-1}$, through the expectations w.r.t. the belief $ \mS_{n-1}$. 
To analyze the  exploration exploitation tradeoff in $\mQ$, we need details of the belief model, which will be explained next. We will come back to the discussion on the properties and behaviors of $\mQ$ in Sec.~\ref{sec:behavior}.
\egroup

%%%%%%%%%%%

%!TEX root = main.tex
\subsection{Belief Model}
\label{sec:belief}
In this section, we briefly describe the Bayesian belief model we use, and explain how Eq.~\ref{eq:acq} is computed with the posterior distribution. The belief model captures our current knowledge of the training curves 
to predict future $\nu^k_r$, and gets updated as new observations become available. 

Our proposed algorithm can work with any Bayesian beliefs which model the training curves\eat{, for example Gaussian Process (GP) and Bayesian neural network}. In this paper, we adopt the Freeze-Thaw GP~\citep{swersky2014freeze}. We use $(k,t)$ to index the hyper-parameters and epochs respectively, and the loss of $k$ from the $t$-th epoch is $y^k(t)$.
The joint distribution of losses from all configurations and epochs, $\vy = [\e^{1}(1), \e^{1}(2), \ldots, \e^{1}(n_1),\ldots, \e^{K}(n_K)]\T$ (arm $k$ has $n_k$ epochs/losses), is given by
\footnote{Assume $N = \sum_{k} n_k$. The dimensionality of variables are: 
{
$\vy \in \R^N, \mK_x \in \R_+^{K\times K}, \mO \in \{0, 1\}^{N\times K}, \ \text{and} \ \mK_{\vt} \in \R_+^{N\times N}$.
}
}
\begin{align}
\Pr (\vy | (\vct{\m},\vt))  = \mathcal{N}
\left(
\vy ;
\vct{0} ,
\mK_{\vct{t}} + \mO \mK_x \mO\T
\right).
\label{eq:GP}
\end{align}
$\mO = \mathrm{block} \diag(\bm{1}_{n_1},  \ldots, \bm{1}_{n_K})$ is a block diagonal matrix of  vector ones. 
Kernel $\mK_x$ models the correlation of asymptote losses across \emph{different configurations}, while kernel
$\mK_{\vct{t}}$ characterizes the correlation of losses from \emph{different epochs} of the same configuration. The entry in $\mK_{\vct{t}}$ is computed via a specific Freeze-Thaw kernel, to capture the decay of losses versus time. 
%   given by $k(t, t') = \frac{\beta^\alpha}{(t+t'+\beta)^{\alpha}}$. It is

We can use the joint distribution (Eq.~\ref{eq:GP}) to predict future performances of $y^{\m}(t), \ \forall k, t$, and update the posteriors with new observations, by applying Bayes' rule. Details of the belief model can be found in Appendix~\ref{app:freezethaw}.

\bgroup
\everymath{\displaystyle}
\paragraph{Computing $\mQ$}
Note that the future best loss is a random variable, $\nu_r^k = \min_{1\leq t \leq r} y^{k}(t_0+t)$.
However, $\nu_r^k$'s distribution is non-trivial to compute, as it is the minimum of $r$ correlated  Gaussians.
\eat{Recall that to compute the action-value function, for every $k$ we need $\nu_r^k = \min_{1\leq t \leq r} y^{k}(t_0+t)$, the future best loss in $r$ epochs staring from some initial epoch $t_0$.  }
To simplify the computation\footnote{We can always use a Monte Carlo approximation to compute this quantity. Asymptotically our simplification does not affect the behavior of the $\mQ$, see Sect.~\ref{sec:behavior}.}, we approximate:
$\nu_r^k \approx y^{k}(t_0+\tau^k)$ where we fix the time index $\tau^k$ deterministically, to be the one which achieves the minimum loss in expectation, $\tau^k = \argmin_{1\leq t \leq r} \expect{}{y^{k}(t_0+t)}$.
The intuition is that for different random draws of the curve, $\nu_r^k$ can be any one of $y^{k}(t_0+t)$ for $1\leq t \leq r$. But with high probability, $\Pr \left[\nu_r^k = y^{k}(t_0+\tau^k) \right] \geq \Pr \left[\nu_r^k = y^{k}(t_0+t) \right], \forall t$.
Thus we use the Gaussian $y^{k}(t_0+\tau^k)$ as $\nu_r^k$. The advantage is that $\mQ_r[a]$ can be computed efficiently in closed-form:
\begin{align}
\mQ_r[a] & = \expect{\nu}{\min\{\nu_r^a, \mu\}} = \mu - \std{}{\nu_r^a} \big(s\Phi(s) +  \phi(s)  \big) 
\label{eq:compute}
\end{align}
where $s = \frac{\mu - \expect{}{\nu_r^a}}{\std{}{\nu_r^a}}$ is the normalized
distance of $\nu_r^a$ to $\mu$, and $\mu$ is a constant, either $\mu_r^\text{1st}$ or $\mu_r^\text{2nd}$ depending on whichever $a$ we are looking at. $\std{}{\cdot}$ is the standard deviation, and $\Phi(\cdot)$ and $\phi(\cdot)$ are the cdf and pdf of standard Gaussian respectively.

\eat{\footnote{$\mu_r^k$ is the expectation of $\nu_r^k$ as defined earlier.}

\eat{\footnote{It is true for independent Gaussian variables. \TODO{In our case, they are positively correlated as $k(t,t') > 0$. CHECK!!!} }}

}
\egroup

%$ \mQ^{(1)}_n$ is computable as $\theta$ is a constant.
%$\mQ^{(2)}_n$ is  computable with the Bayesian posterior distributions because both $\nu^a_{|n}$ and $\nu^{*}_{|n}$ are Gaussian r.v. in our GP belief model (Sec.~\ref{sect:bayesian}).

%!TEX root = main.tex
\subsection{Behavior of $\mQ$}
\label{sec:behavior}
In this section, we analyze the behaviors of the proposed action-value function. 
With Gaussian distributed $\nu_r^k$s, there are nice properties of the proposed $\mQ$. We provide asymptotic analysis of the behaviors as the budget goes to infinity, nonetheless it sheds lights on the behaviors when applied under finite budget. Finite time analysis is more challenging, and is left as future work.

% We provide analysis on the infinite budget sce- nario

When $r$ goes to infinity, $\nu_{\infty}^k$ is the asymptote loss of $k$. Besides, the variance update of $\nu_{\infty}^k$ given by GP, always decreases with more observations, and is independent of the observed values. This leads to the following three properties.

First of all, all arms will be picked infinitely often under $\mQ$. Note the surprise area/VoI for any $k$ (the second term of the r.h.s. in Eq.~\ref{eq:Qsub} and~\ref{eq:Qopt}) is always greater than 0 for any finite time, and approaches 0 in the limit. Hence an arm, which has not been picked for a while, will have higher VoI relative to the rest (since the VoI of other arms go to 0), and get selected again. 
Therefore $\mQ$ is asymptotically consistent: the
best arm will be discovered as the budget goes to infinity, and the objective Eq.~\ref{eq:regret} has
\begin{align*}
\lim_{B\to \infty} \ell^\pi_B - \ell^*_B \to 0.
\end{align*}
Secondly, $\mQ$ balances the exploration with exploitation, since  both configurations with smaller mean (exploitation), or larger variance (exploration) are preferred---in either case the surprise area is large (2nd term on the r.h.s. of Eq.~\ref{eq:Qsub} and~\ref{eq:Qopt}). 
Thirdly, the limiting ratio of the number of pulls between the top and the second best arm approaches 1 assuming the same observation noise parameter. And the limiting ratio between any pair of suboptimal arms is a function of the sub-optimality gap~\citep{ryzhov2016convergence}.
%$\mQ$ with infinite budget coincides to a hybrid EI policy in optimal learning literature as described in, where interested reader is referred to for rigorous proof and details therein.

%When $r \to \infty$, this action-value function leads to several desirable behaviors in terms of how it balances exploration exploitation tradeoff .
%
%First of all, when $\nu_r^k$ follows Gaussian distribution, as given by our belief model, all configurations get to selected infinitely often as the budget $B$ approaches infinity. This guarantees that we will discover the true optimal configuration given unlimited budget.
%
%Secondly, we 

%!TEX root = main.tex

\subsection{Practical Budgeted Hyper-parameter Tuning Algorithm}
\label{sec:alg}
\bgroup
\everymath{\displaystyle}
In this section, we discuss the practical use of $\mQ$ in the budgeted tuning algorithm.

Imagine the behavior of $\mQ$ when applied to the hyper-parameter tuning.  Intuitively, all configurations will get selected often at the beginning, due to the high uncertainty. Gradually as our curve estimation gets more accurate with smaller uncertainty, we will focus the actions on a few promising configurations with good future losses (small sub-optimality gap). Finally, $\mQ$ will mostly allocate budget among the top two configurations. This in practice can be a waste of resource, because we aim to finalize on one model, instead of distinguishing the top two. We propose the following heuristics to fix it.

%some issues of $\mQ$ and propose simple heuristics to fix it. Finally we present the budgeted tuning algorithm. 
\eat{As explained in Sect.~\ref{sec:behavior}, $\mQ$ will mostly allocate budget among the top two arms as the belief model converges. This can be a waste of resource, especially when we almost run out of the budget.  }
\paragraph{Budget Exhaustion}
Recall $\widehat{c}$ is the predicted best arm, and define $\tau^\star = \argmin_{1\leq t \leq r} \expect{}{y^{\widehat{c}}(t_0+t)}$, the number of epoch configuration $\widehat{c}$ short from convergence, if we expect $\widehat{c}$ to hit its minimum. We propose to check the condition $\tau^\star< r$, where $r$ is the remaining budget, to keep track of the budget exhaustion. If it is false, we will pick $\widehat{c}$ ($\cst{else}$ statement in Alg.~\ref{alg}). Note that as we still update our belief after the new observation, we can switch back to the selection rule $\mQ$ ({$\cst{if}$} statement in Alg.~\ref{alg}).

\paragraph{$\varepsilon$-Greedy with Confident Top Arm}
Another drawback of $\mQ$ is that when we have a large number of $K$ (arms), $\mQ$ pulls the top arm less frequently, because the suboptimal arms aggregately could take up a considerable amount of the budget. In fact, we might want to cut down the exploration when we are confident of the top arm. Thus we design the following action selection rule: with probability $\varepsilon$, we select $\widehat{c}$, and follow $\mQ_r[a] = \ProbOpr{E}\left[\min \{ \nu^a_{r},  \mu_{r}^{\text{1st}}\}\right]$ to select the sub-optimal ones for the rest of the time.
\begin{align}
a_n   = \pi^\varepsilon(\xi_{n-1}) = 
    \begin{cases}
  \argmin_{a \neq \widehat{c}}  \mQ_r[a \given \mS_{n-1}]
  \eat{\ProbOpr{E}\left[\min \{ \nu^a_{r},  \mu_{r}^{\text{1st}}\}\right] }, &   \text{ w.p. } \varepsilon,  
     \\
  \widehat{c},   &  \text{ otherwise}. 
    \end{cases}
    \label{eq:policy2}
\end{align} 
We set $\varepsilon=0.5$ in the algorithm. Despite we do not tune $\varepsilon$, $\pi^\varepsilon$ performs well as demonstrated in the experiments.  
The algorithm is summarized in Alg.~\ref{alg}. We will refer to the proposed algorithm as BHPT, and BHPT-$\varepsilon$ in the experiment section.
%\vspace{-8pt}
\begin{algorithm2e}[hbpt]
%\DontPrintSemicolon
\caption{Budgeted Hyperparameter Tuning (BHPT)}\label{alg}
{\bf Input:} Budget $B$, and configurations $[K]$. \\
\For{$n =1, 2, \ldots B$} {
	% Gather performances |_{\sum_{m=1}^Kt_n^{\m} = n}
	\If{$\tau^{\star} < r$}
	{		${a_n = \pi(\xi_{n-1})}$, or ${a_n = \pi^\varepsilon(\xi_{n-1})}$
%		${a_n = \argmin\limits_{a \in [K]} \mQ_r[a | \mS_{n-1}].}$
		\tcp*{Eq.~\ref{eq:policy1} and ~\ref{eq:acq}, or~\ref{eq:policy2}}
	}
	\Else
	{		$a_n =\widehat{c}.$ }%\tcp*{passive mode}
	Run $a_n$ and obtain loss $z_n = \e^{a_n}(t)$ (for some $t$). \\
  Use $z_n$ to update the belief $\mS_n$, $\tau^{\star}$ and $r$. \tcp*{Sect.~\ref{sec:belief} and Appendix}
}
\textbf{Output} $\displaystyle \min_{1\leq n \leq B} z_n$.
\end{algorithm2e}
%\vspace{-10pt}
%\begin{algorithm2e}[hbpt]
%%\DontPrintSemicolon
%\caption{Budgeted Hyperparameter Tuning}\label{alg}
%{\bf Input:} Budget $B$, configurations $[K]$, and $\varepsilon = \{0, 0.5\}$. \\
%\For{$n =1, 2, \ldots B$} {
%	% Gather performances |_{\sum_{m=1}^Kt_n^{\m} = n}
%%	{$\theta \sim \cst{Bernoulli}(\varepsilon)$}
%	\If{$\tau^{\star} < r$}
%	{
%%		\If{$\varepsilon = 0$}{
%			${a_n = \argmin\limits_{a \in [K]} \mQ_r[a | \mS_{n-1}].}$	%\tcp*{active mode}
%		}
%		\ElseIf{$\theta = 1$}{
%			$a_n = \argmin\limits_{a \neq \widehat{c}} \mQ_r[a | \mS_{n-1}].$
%		}
%		\Else{
%			${a_n = \widehat{c}}$
%		}
%	}
%	\Else
%	{
%		$a_n =\widehat{c}.$ %\tcp*{passive mode}
%	}
%	Run $a_n$ and obtain loss $z_n = \e^{a_n}(t)$ (for some $t$). \\
%  Use $z_n$ to update the belief $\mS_n$, $\tau^{\star}$ and $r = B-n$.
%}
%\textbf{Output} $\displaystyle \min_{1\leq n \leq B} z_n$.
%\end{algorithm2e}

\egroup

%!TEX root = main.tex
%\subsection{Bayes Optimal Solution}
\subsection{Discussion}
\label{sec:dp}
\bgroup
\everymath{\displaystyle}

In this section, we discuss and compare with the Bayes optimal solution to the budgeted-tuning problem. The Bayes optimal solution~\cite{ghavamzadeh2015bayesian} handles the challenge of simultaneous estimation the unknown curves and optimization over the future horizon, by solving the following objective,
\begin{align}
\min_{\pi}\expect{}{\ell^\pi_B \given \mS_{0}} = \min_{\pi} \expect{}{\min_{1\leq n \leq B} {z_n} \given[\Big] \mS_{0}} \label{eq:sobjBayes},
\end{align}
where $\mS_0$ is our prior belief over the unknown curves.

To start with, Eq.~\ref{eq:sobjBayes} tries to minimize the loss  \emph{average} across all curves from the prior. 
On the contrary, we try to find the optimal solution $\ell_B^*$ w.r.t. a fixed set of unknown curves. 
Besides the conceptual difference, computationally Bayes optimal solution is taxing due to the exponential growth in the search space~\cite{guez2012efficient}, as it considers subsequent belief updates.
The reduction from the combinatorial to a constant as in our approach, is no longer applicable,
because the optimal planning (Sect.~\ref{sec:structure}) does not hold for Eq.~\ref{eq:sobjBayes}.
%The computation burden of the Bayes optimal solution arise from that they consider the uncertainty of $z_n$ at every step. Instead, our solution only considers the uncertainty of each arm altogether in one random variable $\nu^k_r$.

\paragraph{Rollout}
There is abundant literature of efficient approximate solutions to Eq.~\ref{eq:sobjBayes} in dynamic programming (DP)~\citep{bertsekas1995dynamic}.
We briefly sketch the rollout method in~\citep{lam2016bayesian}, which is the most applicable to our task.  It applies the one-step lookahead technique, with two approximations to compute the future rewards: first it truncates the horizon of belief updates to at most $h$ rolling steps, where $h$ is a  parameter. 
Secondly, they propose to use expected improvement (\EI) from the \BO literature as the sub-optimal rollout heuristics to collect the future rewards.  Details of the rollout algorithm can be found in the Appendix~\ref{app:dp}.

As we will see in the experiment section, the truncated horizon and the lack of long-term prediction is detrimental for the tuning performance. 
On the contrary, predicting $\nu_r^k$ directly in our approach is both computationally efficient and conceptually advantageous for the hyper-parameter tuning problem.

\egroup

%% useless
%The action-value function is given by
%% \hspace*{-10cm} 
%{
%\begin{align}
%&H_h(\widetilde{\mS}_{n+h-1})  = \max_a \ProbOpr{E} \left[(\zeta_{n+h-1} - \nu_{r-h}^a)^+ \given[\Big] \widetilde{\mS}_{n+h-1} \right], \label{eq:rolloutT}\\
%&H_k(\widetilde{\mS}_{n+k-1})  = \max_a \ProbOpr{E} \left[(\zeta_{n+k-1} - z_{n+k-1}^{a})^+ \given[\Big] \widetilde{\mS}_{n+k-1} \right] \notag \\
%&+ H_{k+1}(\widetilde{\mS}_{n+k}), \text{ for } k=1, \ldots, h-1. \label{eq:rollout}
%\end{align}
%}
%Note both Eq.~\ref{eq:rolloutT} and Eq.~\ref{eq:rollout} select the rollout action $a$ based on a greedy heuristic.

%!TEX root = main.tex
\section{Related Work}
\label{sec:related}
% hyper-parameter optimization
Automated design of machine learning system is an important research topic~\cite{snoek2012practical}. Traditionally, hyper-parameter optimization (HO) is formulated as a black-box optimization and solved by Bayesian optimization (\BO). It uses a probabilistic model together with an acquisition function, to adaptively select configurations to train in order to identify the optimal one~\citep{shahriari2016taking}.  However oftentimes, fully training a single configuration can be expensive. Therefore recent advances focus to exploit cheaper surrogate tasks to speed up the tuning. For example, Fabolas~\citep{klein2017fast}  evaluates configurations on subsets of the data and extrapolates on the whole set;  FreezeThaw~\citep{swersky2014freeze} uses the partial training curve to predict the final performance.

Recently Hyperband~\cite{li2016hyperband} formulates the HO as a non-stochastic best-arm identification problem. It proposes to adaptively evaluate a configuration's intermediate results and quickly eliminate poor performing arms. In a latest work, ~\citep{falkner2018bohb} combines the benefits of \BO and \HB to achieve fast convergence to the optimal configuration. Other HO work includes gradient-based approach~\citep{maclaurin2015gradient, franceschi2017forward}, meta-learning~\citep{jaderberg2017population,franceschi2018bilevel}, and the spectral approach~\cite{hazan2017hyperparameter}. Please also refer to Table~\ref{table:compare} in Sect.~\ref{sec:exp} for a comparison.

While all these works improve the efficiency of hyper-parameter tuning for large-scale learning systems, none of them has an explicit notion of (hard) budget constraint for the tuning process. Neither do they consider to adapt the tuning strategy across different budgets. On the contrary, we propose to take the resource constraint as an input to the tuning algorithm, and balance the exploration and exploitation tradeoff w.r.t. a specific budget constraint. 

Table.~\ref{table:compare} summarizes the comparisons of popular tuning algorithms from three perspectives: whether it uses a (probabilistic) model to adaptively predict and identify good configurations;  whether it supports early stop and resume; and whether it adapts to different budgets. 

\vspace{-5pt}
\begin{table}[hbtp]
\centering
\caption{Comparison of tuning algorithms}
\label{table:compare}
%\resizebox{0.49\textwidth}{!}{ 
\begin{tabular}{r|c|c|c}
\hline
%\multirow{2}{*}{algorithm} & early stop & adaptive (future) & budget  \\ 
%& and resume & prediction & aware \\ \hline
algorithm & early stop and resume & adaptive (future) prediction& budget  aware\\ \hline
random search &  $\times$ & $\times$ & $\times$\\ 
\BO (GP-EI/SMAC) & $\times$ &  \checkmark & $\times$\\
 Fabolas & \checkmark &  \checkmark & $\times$\\
 FreezeThaw & \checkmark & \checkmark & $\times$\\
 \HB &  \checkmark  & $\times$ & $\times$\\ %\kern-1.1ex\raisebox{.7ex}{\rotatebox[origin=c]{125}{--}}\footnotemark
 Rollout & \checkmark &  \checkmark &  \checkmark \\ \hline %\kern-1.1ex\raisebox{.7ex}{\rotatebox[origin=c]{125}{--}}\footnotemark 
BHPT, BHPT-$\varepsilon$&  \checkmark &  \checkmark &  \checkmark\\
 \hline
\end{tabular}
%}
\end{table}

% budgeted optimization
There are other budgeted optimization formulations solving problems in different domains. ~\cite{lam2016bayesian} proposed an approximate dynamic programming approach (Rollout) to solve \BO under a finite budget, see Sect.~\ref{sec:dp} and~\ref{sec:exp} for discussions and comparisons. 
For the budget optimization in online advertising, ~\cite{boutilier2016budget} studies the constrained MDP~\citep{sun2008constrained}, where in a Markov Decision Process (MDP) each action also incurs some cost, and the policy should satisfy constraints on the total cost. However, unlike in advertising that the cost of different actions can vary or even be random, in hyper-parameter tuning the human operator usually can pre-specify and determine the cost (of actions) -- train the configuration for an epoch and stop. Thus, a finite-horizon formulation is simpler and more suitable for the tuning task compared to the constrained formulation.

The sequential decision making problem introduced in our paper is an instance of extreme bandits~\citep{nishihara2016no}. A known challenge in extreme bandits is that the optimal policy for the remaining horizon depends on the past history, which implies that the policy should not be state-less. On the other hand, if we formulate the hyper-parameter tuning as a MDP, we suffer from that the states are never revisited. Existing approaches in MDP~\citep{ghavamzadeh2015bayesian} can not be directly applied.

\section{Experiment}
\label{sec:exp}
In this section, we first  compare and validate the conceptual advantage of the BHPT algorithm over other methods on synthetic data, by analyzing the exploration exploitation tradeoff under different budgets. Then we demonstrate the performance of the BHPT algorithm on real-world hyper-parameter tuning tasks.
Particularly, we include a task to tune network architectures because selecting the optimal architecture  under a budget constraint is of great practical importance.
\eat{We choose 4 real-world tasks, including classification for supervised learning and generative model for unsupervised learning. }

We start by describing the experimental setups, \ie the data, evaluation metric and baseline methods, in Sec.~\ref{sect:exp_descrip}, and then provide results and analysis in Sec.~\ref{sect:syn} and~\ref{sect:real}. 

\subsection{Experiment Description}
\label{sect:exp_descrip}
\paragraph{Data and Evaluation Metric}

\begin{table}[hbtp]
\centering
\caption{Data and Evaluation Metric}\label{table:data}
%\resizebox{0.5\textwidth}{!}{ 
\begin{tabular}{l|c|c|c}
\hline
%\multirow{2}{*}{dataset} & $K$  & \# hyper- & \multirow{2}{*}{evaluation} \\ 
%& \# arms & params. &   \\ \hline
dataset & \# arms $K$  & \# hyper-parameters & evaluation \\ \hline
%dataset & \shortstack[c]{$K$ \\ \# arms} & \shortstack[c]{\# hyper-\\ params} & evaluation \\ \hline
synthetic & 84 & NA & Eq.~\ref{eq:gap}  \\ \hline
ResNet on CIFAR-10 &  96  &  5 & error rate \\
FCNet on MNIST &  50  &  10 & error rate \\
VAE on MNIST &  49  &  4 & ELBO \\
ResNet/AlexNet on CIFAR-10&  49  &  6 & error rate \\ \hline
\end{tabular}
%}
\end{table}
For the data preparation, we generate  and save the learning curves of all configurations, to avoid repeated training when tuning under different budget constraints.
For synthetic set, we generate 100 sets of training curves drawn from a Freeze-Thaw GP. 
For real-world set, we create 4 tuning tasks as summarized in Table~\ref{table:data}. For more details, please refer to the Appendix~\ref{app:exp}.

For evaluation metric on synthetic data, we define the \emph{normalized regret}
\begin{equation}
\mathcal{R}_B = \frac{\ell^\pi_B -\ell^{*}_B}{\ell_0 - \ell^{*}_B}, \label{eq:gap}
\end{equation}
where $\ell_0$ is the initial loss of all arms; $\ell^\pi_B$ is the tuning output; $\ell^{*}_B$ is the optimal solution with known curves. We normalize the regret over the data range $\ell_0 - \ell^{*}_B$.
For real-world task, see Table~\ref{table:data} for the evaluation metrics.
As all tasks are loss minimization problems, the reported measure  is the smaller the better.

\paragraph{Baseline Methods}
We compare to Hyperband,  Bayesian Optimization (BO) and its variants (Fabolas and FreezeThaw), as well as the rollout solution described in Sect.~\ref{sec:dp}.
For descriptions and comparisons see Sect.~\ref{sec:related}, and refer to Table~\ref{table:compare} for a summary. 
Implementation details can be found in Appendix~\ref{app:exp}.

%Details see Sect.~\ref{app:dp} and~\cite{lam2016bayesian}.

% \theadfont  

%\addtocounter{footnote}{-2}
%\stepcounter{footnote}\footnotetext{\HB does not support resume.}
%\stepcounter{footnote}\footnotetext{In rollout, the rolling horizon truncates the planning path, which makes it only partially adaptive to the budget.}

%\HB early stops but cannot resume a eliminated configuration. 
%In rollout algorithm, even though the budget constraint is input to the algorithm, the computation burden limits the planning to the truncated rolling horizon, which is not adaptive to the true budget.

%%%%%%%%%%
% !TEX root = main.tex

\subsection{Results on Synthetic Data}
\label{sect:syn}
% change place
\begin {figure*}[!hbtp]%[!hbtp]
\centering
\begin{adjustbox}{width=\textwidth}
\centering
%\hspace*{-1cm}
\begin{tabular}{ccc}
\multicolumn{3}{c}{\includegraphics[clip=true, width=1.05\textwidth]{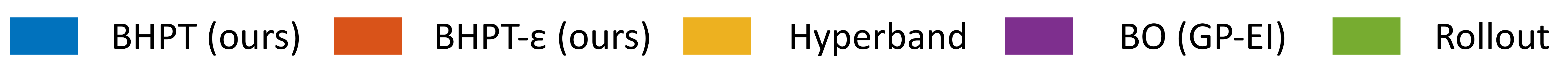}} \\
\includegraphics[width=0.4\textwidth]{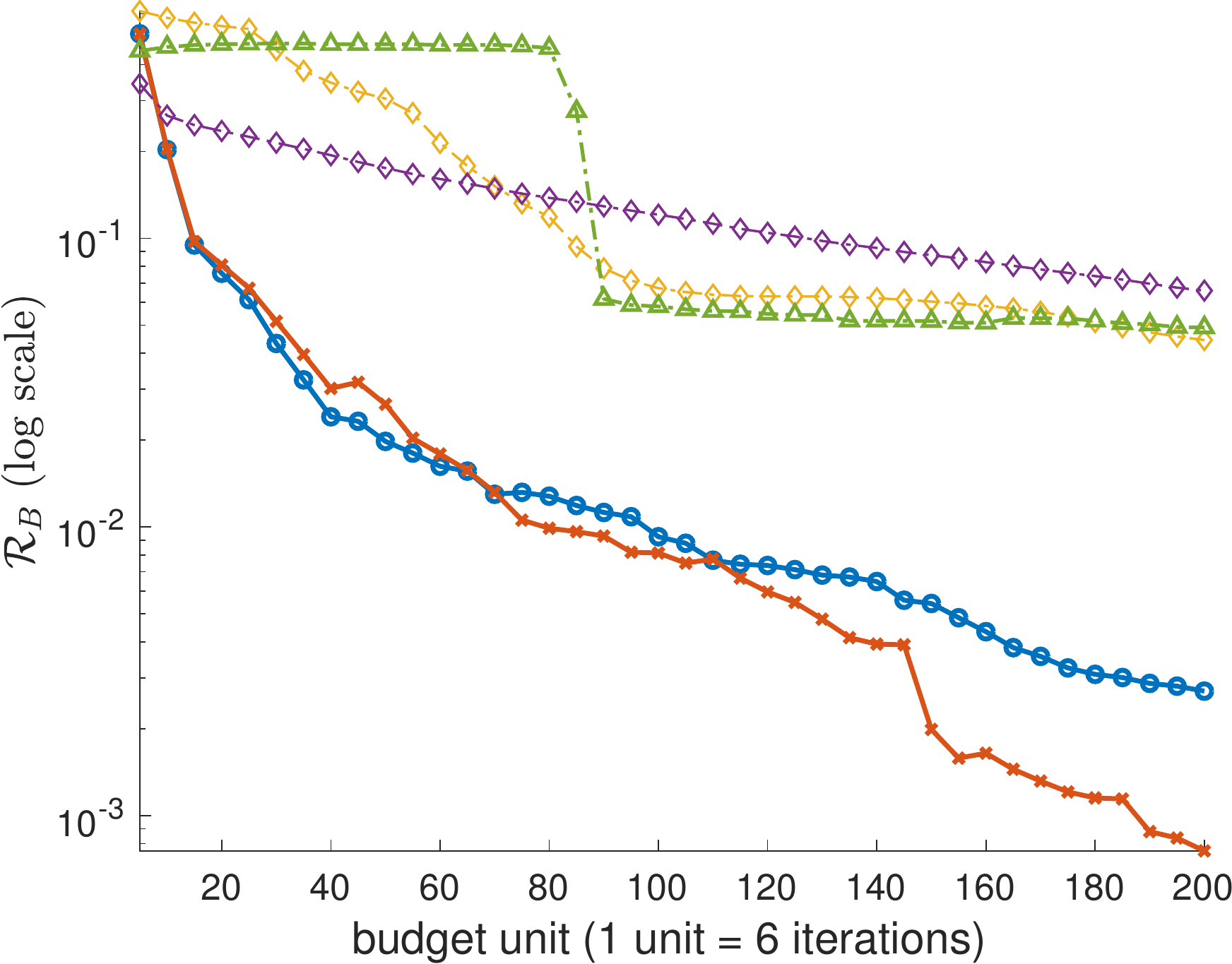}
&
\includegraphics[width=0.4\textwidth]{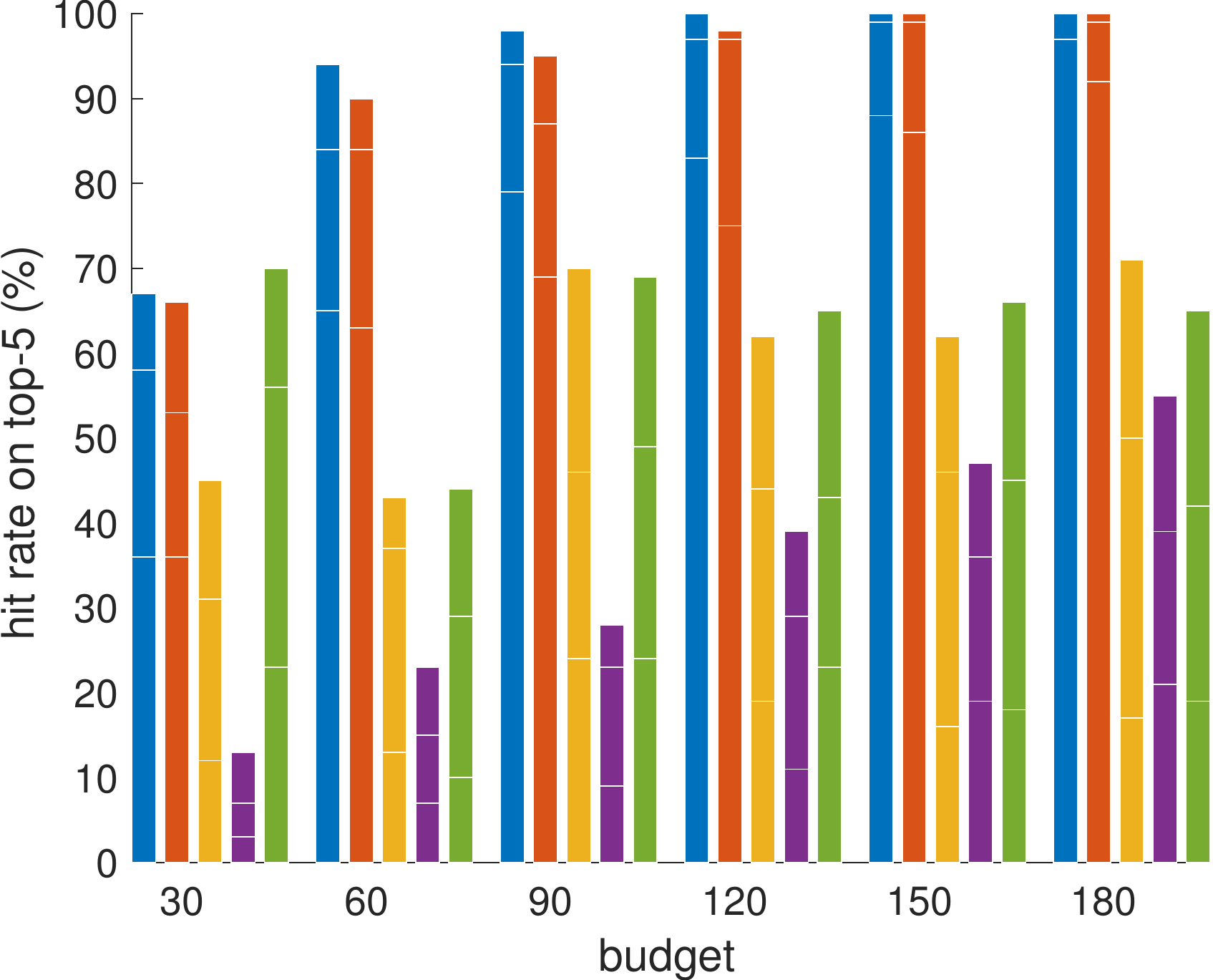} % arm_in_top5
&
 \includegraphics[width=0.4\textwidth]{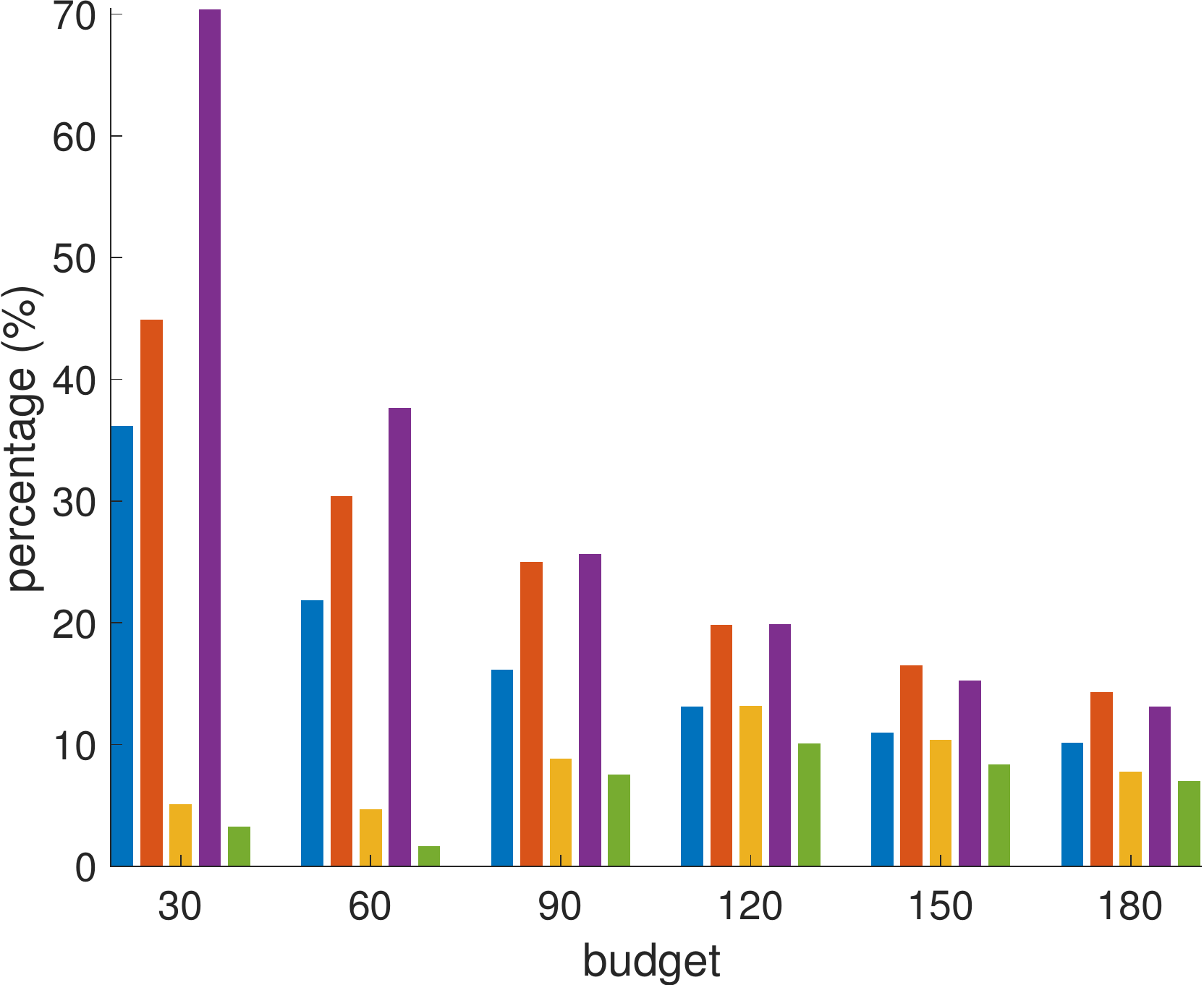}  
\\
 {(a)  normalized regret $\mathcal{R}_B$ }
&{ (b) exploration: output arm hit rate @ top 5 }
& {(c) exploitation: \% budget on the output arm} \\
\small{\thead{For BHPT, $\mathcal{R}_B \to 0$ as $B$ increases.
\BO is better under  \\ small  $B$,
while   Hyperband is better under large $B$.}} 
& 
\small{\thead{The hit rate of \BO and our methods \\
improves as $B$ increases.}}
& 
\small{\thead{The percentage of budget on the output arm of \\
\BO and our methods decreases as $B$ increases.}}
\end{tabular}
\end{adjustbox}
\caption{Budgeted Optimization on 100 Synthetic Sets.}
\label{syn}
\end{figure*}

On synthetic data, because we use the correct prior, we expect that the belief model  learns to accurately predict the future as the budget increases. Thus we can examine the behaviors of the proposed algorithms under different budgets. All results are averaged over 100 synthetic sets. 

First, we plot out the normalized regrets $\mathcal{R}_B$ (Eq.~\ref{eq:gap}) over budgets in Fig.~\ref{syn}(a).
% result description
Our policies consistently outperform competing methods under different budget constraints. 
As the budget increases, the proposed BHPT methods $\mathcal{R}_B \to 0$ as discussed in Sect.~\ref{sec:behavior}.

\paragraph{Exploration vs. Exploitation}
There are two factors that determine the performance of an algorithm: whether it correctly identifies the optimal arm, and whether it spends sufficient budget to achieve small loss on such arm. Loosely speaking, the first task reflects if the algorithm achieves effective exploration, such that it can accurately estimate the curves and identify the top arm, and the second task indicates sufficient exploitation. 
To examine the ``exploration'', we plot the hit rate of the output arm on the top 5 arms (out of all $84$) across different budgets in Fig.~\ref{syn}(b)\footnote{The three stacks in each bar are the hit rate @ top-1, top-3, and top-5 in ground-truth respectively.}.
To check the ``exploitation'', we visualize the percentage of the budget spent on the output arm in Fig.~\ref{syn}(c). In both (b) and (c), the higher of the bar the better.

% compare different methods
In (b), both BHPTs do better in exploration than all baselines. The ``adaptive prediction'' column in Table.~\ref{table:compare} explains the exploration behavior.
The probabilistic prediction module in \BO and our methods improve with more budgets, which explains the increase in hit rate as budget increases.
In (c), our methods perform well in terms of exploitation. 
The ``early stop'' column in Table.~\ref{table:compare} partially explains the exploitation behavior.
\BO does strong exploitation under small budget because it does not early stop configurations, which also explains its (relatively) good performance under small budgets in (a).

Despite the Rollout has a belief model and does future predictions as BHPT, it doesn't perform well on neither task:
the hit rate does not improve with more budget, nor does it exploit sufficiently on the output arm. 
It is because the rollout truncates the planning horizon due to the computation challenge, which leads to myopic behaviors and the poor results. 
Indeed in all three subplots of Fig.~\ref{syn}, Rollout performs  and  behaves similarly to \HB, which only uses current performance to select actions. 
This demonstrates the importance of \emph{long-term} predictions and planning in the budgeted tuning task.

\eat{Rollout performs  as both of them use myopic information to select configurations.} 

% our method across budget
Comparing (b) and (c), there is a clear tradeoff between exploration and exploitation, that the hit rate decreases as the budget percentage increases. 
Note that BHPT adjusts this tradeoff automatically across different budgets. 
Compare the BHPT against its $\varepsilon$-greedy variant, the BHPT-$\varepsilon$ does slightly better in exploitation, and worse in exploration, as expected.

\eat{
\paragraph{Computation} In table~\ref{table:time}, we report the computation cost in seconds for the tuing algorithms under $B=100$. Hyperband is the fastest because it's based on random search. BO updates the Gaussian Process once a full curve is revealed (one fully trained configuration in hyper-parameter tuning), thus it is slightly more expensive. BHPT updates the GP and does re-planning at every step (100 times re-planning in this example), which adds to the computation cost. Rollout is significantly more expensive than our BHPT due to the imaginary rollout and belief updates. The proposed BHPT achieves better tuning performance with a moderate increase in computation cost.

\begin{table}[hbtp]
\centering
\caption{Computation Time}\label{table:time}

\resizebox{0.45\textwidth}{!}{ 
\begin{tabular}{c|c|c|c|c}
\hline
algorithm &  BHPT & Hyperband & BO & Rollout  \\ \hline
time (s) & 11.89 & 0.01 & 0.40 & 1007.06 \\\hline
\end{tabular}
}
%\vspace{-5pt}
\end{table}
}

\subsection{Results on Real-world Data}
\label{sect:real}

\begin {figure*}[!hbtp] %[!hbtp]%[!hbtp]
%\captionsetup{farskip=0pt}% <--- no gap at the top
\centering
%\begin{adjustbox}{width=\textwidth}
\hspace*{-1cm}
\begin{tabular}{ccc}
\multicolumn{3}{c}{\includegraphics[clip=true, width=\textwidth]{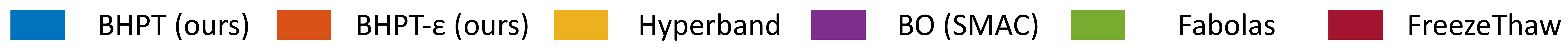}} \\
\includegraphics[width=0.35\textwidth]{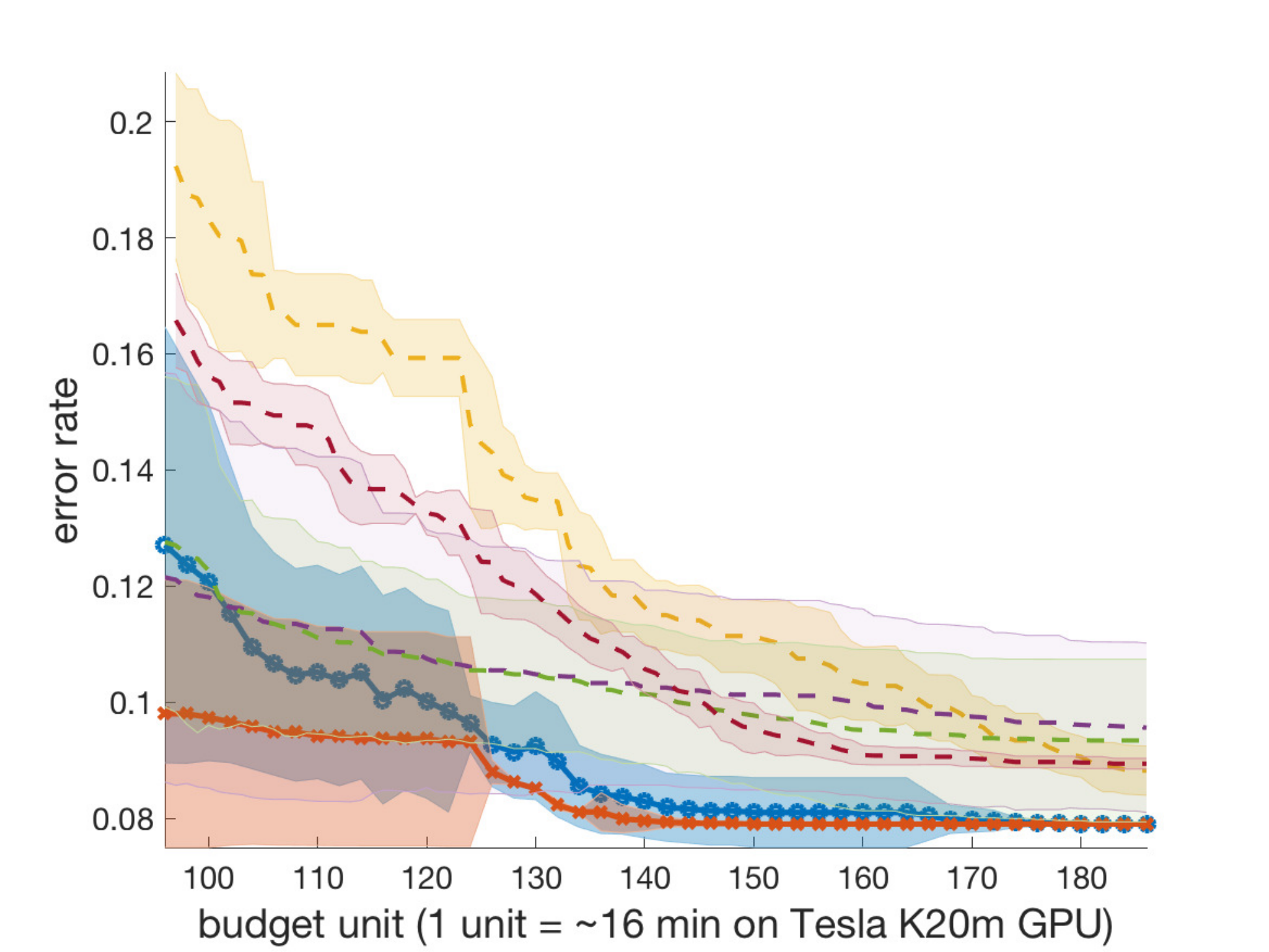} %[width=0.45\textwidth]
& \includegraphics[width=0.35\textwidth]{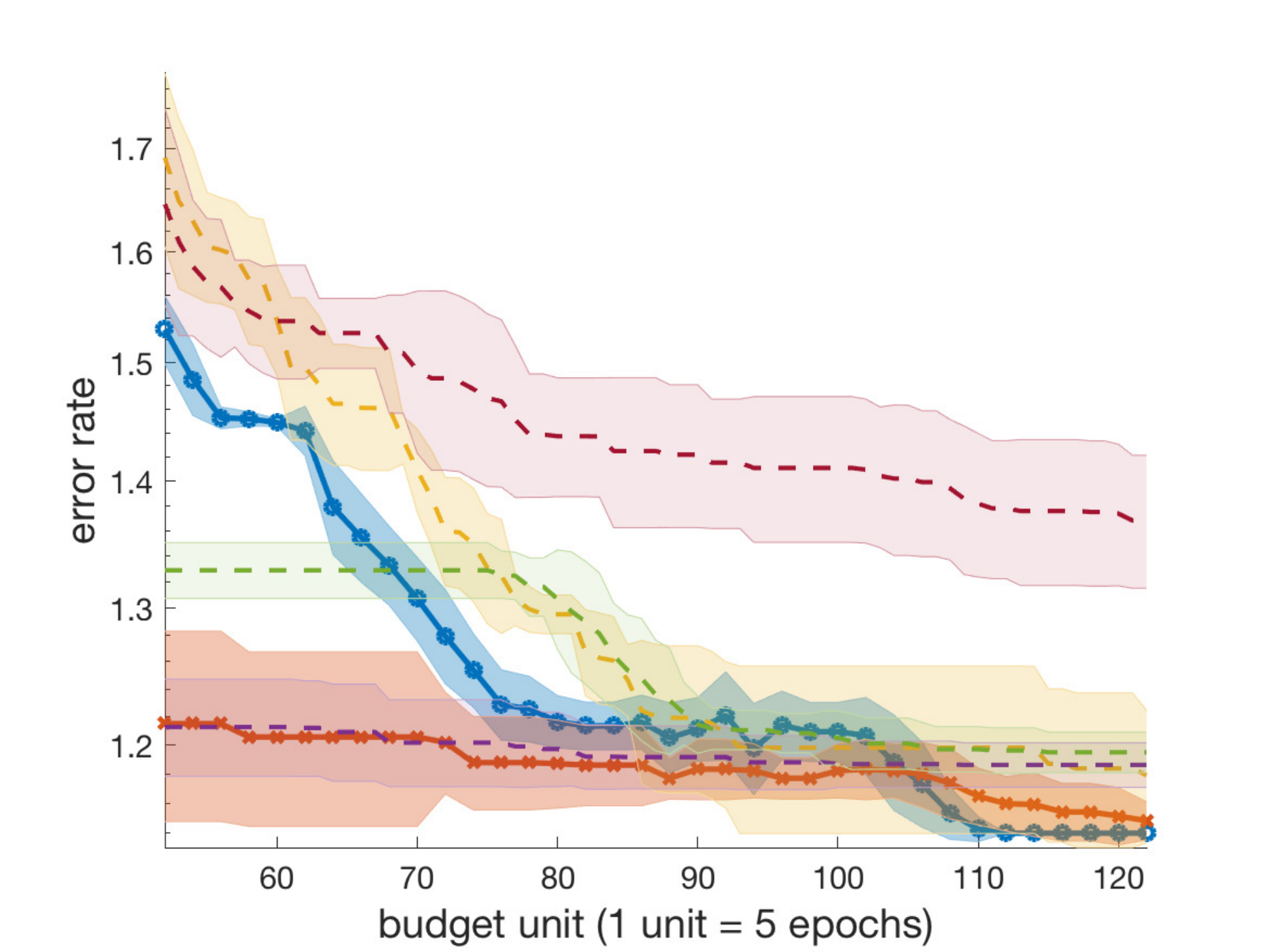}
&\includegraphics[width=0.35\textwidth]{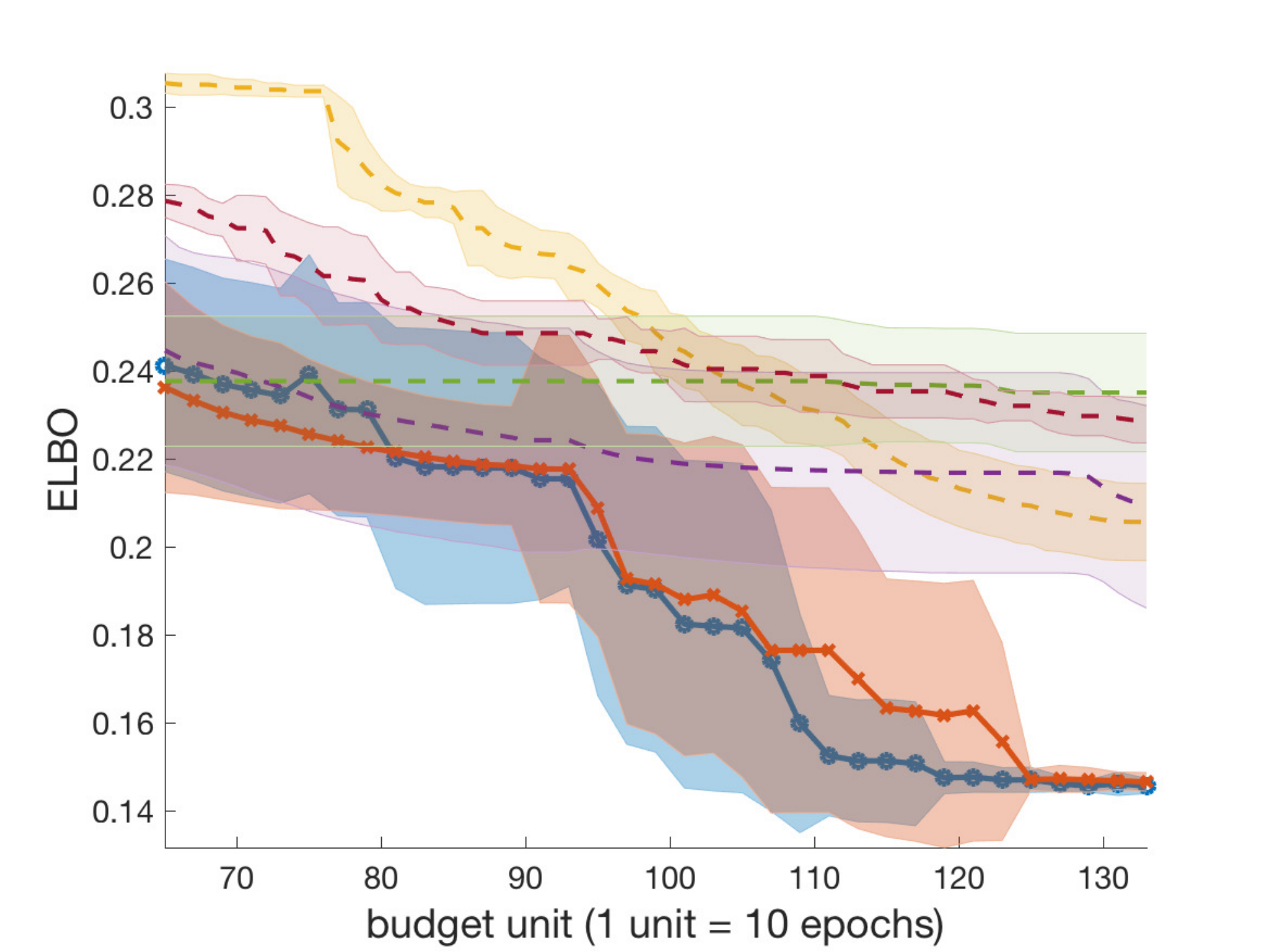}
 \\
 { (a) ResNet for classification}
&
{ (b) FCnet for classification}
& {(c) VAE for generative model} \\
\multicolumn{3}{c}{BHPT (blue)  converges to the global optimal model at the rightmost budget for all tasks. }\\
\multicolumn{3}{c}{BHPT-$\varepsilon$ is better under small budgets, while BHPT is better under large budgets.}
\end{tabular}
\caption{Real-world Hyper-parameter Tuning Tasks}
\label{real}
\end{figure*}

In this section, we report the tuning performances on real-world tuning tasks across different budget constraints.
We plot the tuning outcomes (error rate or ELBO) over budgets in Fig.~\ref{real} and Fig.~\ref{arch}(a). 
Each curve is the average of 10 runs from different random seeds, and the mean with one standard deviation is shown in the figures.

BHPT methods work well under a wide range of budgets, and outperform \BO, \HB, FreezeThaw and the state-of-the-art algorithm Fabolas, across 4 tuning tasks. The trend of different methods across budgets is consistent with the observations on the synthetic data.

As explained in the synthetic data experiment, the vanilla BHPT does better in exploration while the $\varepsilon$-greedy variant does more exploitation. This explains the superior performance of the $\varepsilon$-greedy under small budgets. 
However, the lack of exploration results in worse belief model and damages the performance as the budget increases. 
This phenomenon is more salient on task Fig.~\ref{arch}(a), where the belief modeling is more challenging due to different learning curve patterns between ResNet and AlexNet. 

\paragraph{Budget Adaptive Behaviors}
\begin{figure}[htbp]
\centering
%\hspace*{-.9cm}
\begin{tabular}{cc}
\includegraphics[width=0.45\textwidth]{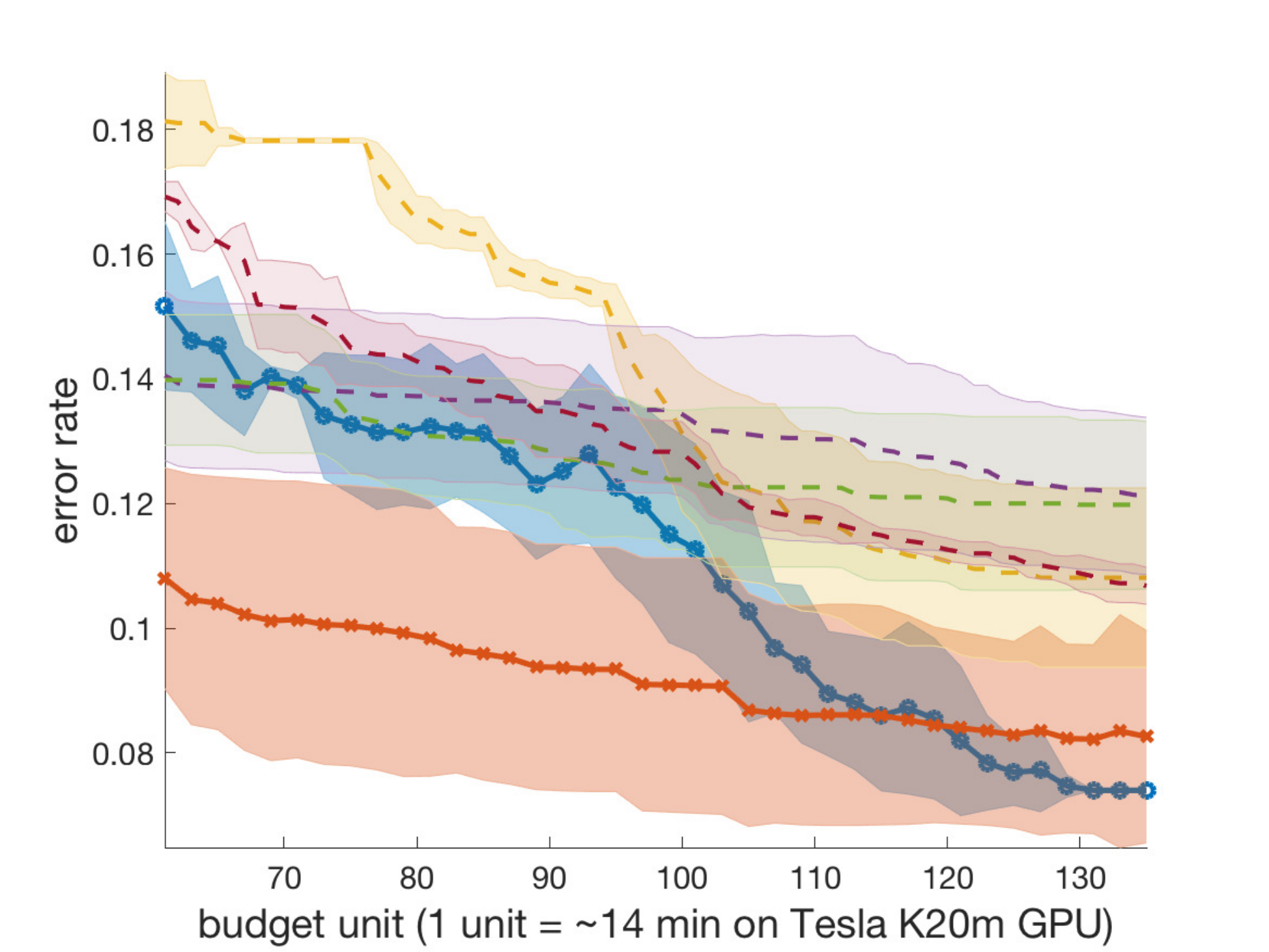}
&
%\hspace*{-0.3in}
 \includegraphics[width=0.41\textwidth]{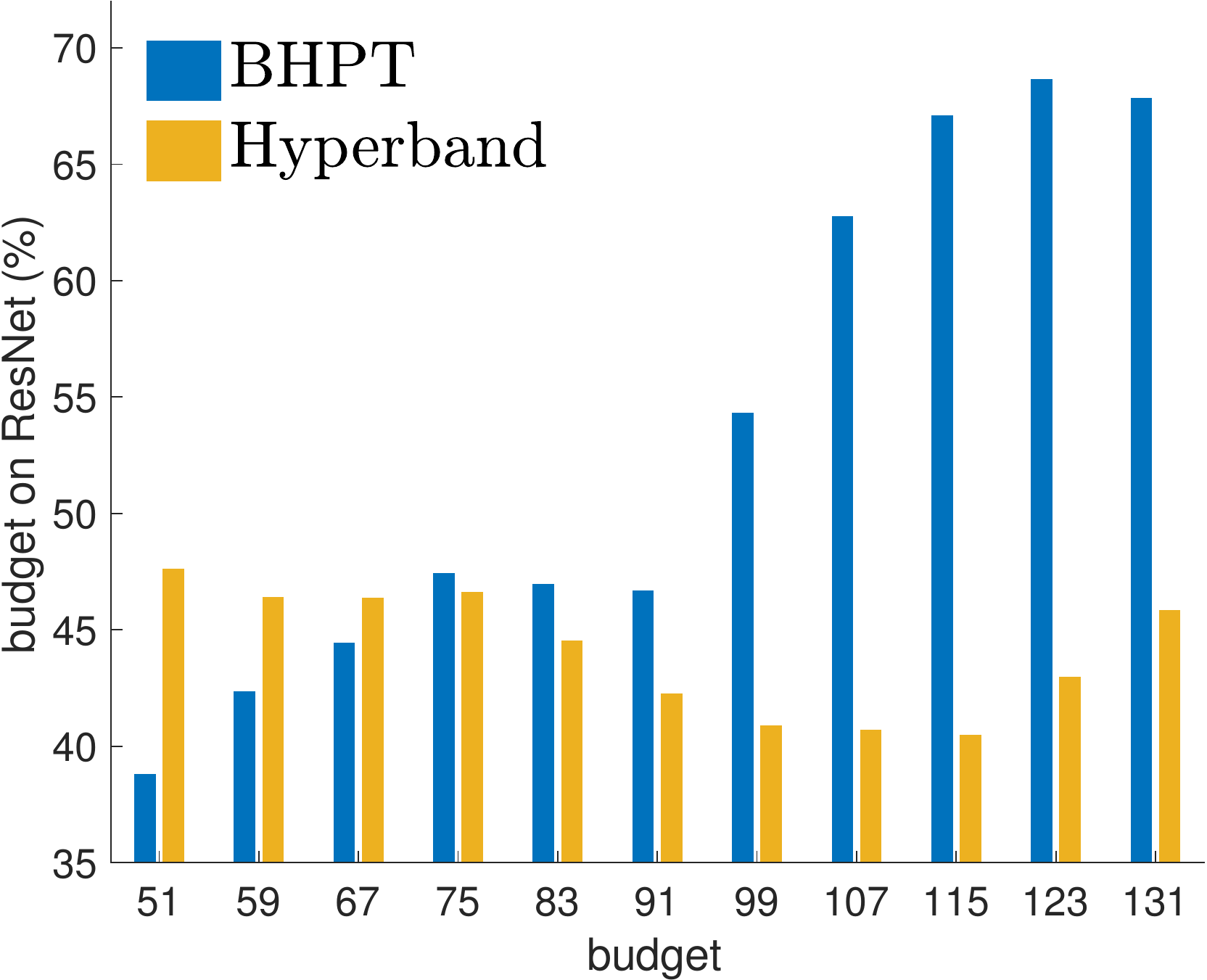} \\
{ (a) AlexNet and ResNet }
& 
%\hspace*{-0.3in}
{(b) budget spent on ResNet}
\end{tabular}
\caption{On an architecture selection task, the proposed BHPT adapts its behavior to different budget constraints, while Hyperband employs a fixed strategy.}
\label{arch}
\end{figure}

An important motivation to study the budgeted tuning problem 
is that it is practically desirable to have adaptive tuning strategy under different constraints. For example, the optimal configuration might change under different budgets.
We would like to examine whether the proposed BHPT exhibits such adaptive behavior. 
We take the architecture selection task between ResNet and AlexNet on \cifar, and  visualize the ratio of the resource spent on ResNet over budgets in Fig.~\ref{arch}(b).
ResNet converges slower than AlexNet, but reaches a smaller error rate. Thus it's rewarding to focus the tuning on AlexNet under small budgets, and vice versa. 
Indeed, the BHPT allocates more resource to ResNet as the budget increases, compared to the baseline Hyperband, which samples the two networks more or less uniformly at random.

\section{Conclusion}
\label{sec:conclusion}
In this paper, we study the budgeted hyper-parameter tuning problem. We formulate a sequential decision making problem, and propose an algorithm, which uses long-term predictions with an action-value function to balance the exploration exploitation tradeoff. It exhibits budget adaptive tuning behavior, and achieves the state-of-the-art performance across different budgets on real-world tuning tasks.

\small{
\bibliographystyle{plainnat}
\bibliography{ref}

\begin{thebibliography}{32}
\providecommand{\natexlab}[1]{#1}
\providecommand{\url}[1]{\texttt{#1}}
\expandafter\ifx\csname urlstyle\endcsname\relax
  \providecommand{\doi}[1]{doi: #1}\else
  \providecommand{\doi}{doi: \begingroup \urlstyle{rm}\Url}\fi

\bibitem[Bello et~al.(2017)Bello, Zoph, Vasudevan, and Le]{bello2017neural}
Irwan Bello, Barret Zoph, Vijay Vasudevan, and Quoc~V Le.
\newblock Neural optimizer search with reinforcement learning.
\newblock \emph{arXiv preprint arXiv:1709.07417}, 2017.

\bibitem[Bertsekas et~al.(1995)Bertsekas, Bertsekas, Bertsekas, and
  Bertsekas]{bertsekas1995dynamic}
Dimitri~P Bertsekas, Dimitri~P Bertsekas, Dimitri~P Bertsekas, and Dimitri~P
  Bertsekas.
\newblock \emph{Dynamic programming and optimal control}, volume~1.
\newblock Athena scientific Belmont, MA, 1995.

\bibitem[Boutilier and Lu(2016)]{boutilier2016budget}
Craig Boutilier and Tyler Lu.
\newblock Budget allocation using weakly coupled, constrained markov decision
  processes.
\newblock In \emph{UAI}, 2016.

\bibitem[Bubeck et~al.(2012)Bubeck, Cesa-Bianchi, et~al.]{bubeck2012regret}
S{\'e}bastien Bubeck, Nicolo Cesa-Bianchi, et~al.
\newblock Regret analysis of stochastic and nonstochastic multi-armed bandit
  problems.
\newblock \emph{Foundations and Trends{\textregistered} in Machine Learning},
  5\penalty0 (1):\penalty0 1--122, 2012.

\bibitem[Chen et~al.(2016)Chen, Hoffman, Colmenarejo, Denil, Lillicrap, and
  de~Freitas]{chen2016learning}
Yutian Chen, Matthew~W Hoffman, Sergio~Gomez Colmenarejo, Misha Denil,
  Timothy~P Lillicrap, and Nando de~Freitas.
\newblock Learning to learn for global optimization of black box functions.
\newblock \emph{arXiv preprint arXiv:1611.03824}, 2016.

\bibitem[Dearden et~al.(1998)Dearden, Friedman, and
  Russell]{dearden1998bayesian}
Richard Dearden, Nir Friedman, and Stuart Russell.
\newblock Bayesian q-learning.
\newblock In \emph{AAAI/IAAI}, pages 761--768, 1998.

\bibitem[Dearden et~al.(1999)Dearden, Friedman, and Andre]{dearden1999model}
Richard Dearden, Nir Friedman, and David Andre.
\newblock Model based bayesian exploration.
\newblock In \emph{Proceedings of the Fifteenth conference on Uncertainty in
  artificial intelligence}, pages 150--159. Morgan Kaufmann Publishers Inc.,
  1999.

\bibitem[Falkner et~al.(2018)Falkner, Klein, and Hutter]{falkner2018bohb}
Stefan Falkner, Aaron Klein, and Frank Hutter.
\newblock Bohb: Robust and efficient hyperparameter optimization at scale.
\newblock \emph{arXiv preprint arXiv:1807.01774}, 2018.

\bibitem[Franceschi et~al.(2017)Franceschi, Donini, Frasconi, and
  Pontil]{franceschi2017forward}
Luca Franceschi, Michele Donini, Paolo Frasconi, and Massimiliano Pontil.
\newblock Forward and reverse gradient-based hyperparameter optimization.
\newblock \emph{arXiv preprint arXiv:1703.01785}, 2017.

\bibitem[Franceschi et~al.(2018)Franceschi, Frasconi, Salzo, and
  Pontil]{franceschi2018bilevel}
Luca Franceschi, Paolo Frasconi, Saverio Salzo, and Massimilano Pontil.
\newblock Bilevel programming for hyperparameter optimization and
  meta-learning.
\newblock \emph{arXiv preprint arXiv:1806.04910}, 2018.

\bibitem[Ghavamzadeh et~al.(2015)Ghavamzadeh, Mannor, Pineau, Tamar,
  et~al.]{ghavamzadeh2015bayesian}
Mohammad Ghavamzadeh, Shie Mannor, Joelle Pineau, Aviv Tamar, et~al.
\newblock Bayesian reinforcement learning: A survey.
\newblock \emph{Foundations and Trends{\textregistered} in Machine Learning},
  8\penalty0 (5-6):\penalty0 359--483, 2015.

\bibitem[Guez et~al.(2012)Guez, Silver, and Dayan]{guez2012efficient}
Arthur Guez, David Silver, and Peter Dayan.
\newblock Efficient bayes-adaptive reinforcement learning using sample-based
  search.
\newblock In \emph{Advances in Neural Information Processing Systems}, pages
  1025--1033, 2012.

\bibitem[Hazan et~al.(2017)Hazan, Klivans, and Yuan]{hazan2017hyperparameter}
Elad Hazan, Adam Klivans, and Yang Yuan.
\newblock Hyperparameter optimization: a spectral approach.
\newblock \emph{arXiv preprint arXiv:1706.00764}, 2017.

\bibitem[He et~al.(2016)He, Zhang, Ren, and Sun]{he2016deep}
Kaiming He, Xiangyu Zhang, Shaoqing Ren, and Jian Sun.
\newblock Deep residual learning for image recognition.
\newblock In \emph{Proceedings of the IEEE Conference on Computer Vision and
  Pattern Recognition}, pages 770--778, 2016.

\bibitem[Howard(1966)]{howard1966information}
Ronald~A Howard.
\newblock Information value theory.
\newblock \emph{IEEE Transactions on systems science and cybernetics},
  2\penalty0 (1):\penalty0 22--26, 1966.

\bibitem[Jaderberg et~al.(2017)Jaderberg, Dalibard, Osindero, Czarnecki,
  Donahue, Razavi, Vinyals, Green, Dunning, Simonyan,
  et~al.]{jaderberg2017population}
Max Jaderberg, Valentin Dalibard, Simon Osindero, Wojciech~M Czarnecki, Jeff
  Donahue, Ali Razavi, Oriol Vinyals, Tim Green, Iain Dunning, Karen Simonyan,
  et~al.
\newblock Population based training of neural networks.
\newblock \emph{arXiv preprint arXiv:1711.09846}, 2017.

\bibitem[Jamieson and Talwalkar(2016)]{jamieson2016non}
Kevin Jamieson and Ameet Talwalkar.
\newblock Non-stochastic best arm identification and hyperparameter
  optimization.
\newblock In \emph{Artificial Intelligence and Statistics}, pages 240--248,
  2016.

\bibitem[Klein et~al.(2017{\natexlab{a}})Klein, Falkner, Bartels, Hennig, and
  Hutter]{klein2017fast}
Aaron Klein, Stefan Falkner, Simon Bartels, Philipp Hennig, and Frank Hutter.
\newblock Fast bayesian optimization of machine learning hyperparameters on
  large datasets.
\newblock In \emph{Artificial Intelligence and Statistics}, pages 528--536,
  2017{\natexlab{a}}.

\bibitem[Klein et~al.(2017{\natexlab{b}})Klein, Falkner, Springenberg, and
  Hutter]{klein2017learning}
Aaron Klein, Stefan Falkner, Jost~Tobias Springenberg, and Frank Hutter.
\newblock Learning curve prediction with bayesian neural networks.
\newblock \emph{Proc. of ICLR}, 17, 2017{\natexlab{b}}.

\bibitem[Krizhevsky and Hinton(2009)]{krizhevsky2009learning}
Alex Krizhevsky and Geoffrey Hinton.
\newblock Learning multiple layers of features from tiny images.
\newblock 2009.

\bibitem[Krizhevsky et~al.(2012)Krizhevsky, Sutskever, and
  Hinton]{krizhevsky2012imagenet}
Alex Krizhevsky, Ilya Sutskever, and Geoffrey~E Hinton.
\newblock Imagenet classification with deep convolutional neural networks.
\newblock In \emph{Advances in neural information processing systems}, pages
  1097--1105, 2012.

\bibitem[Lam et~al.(2016)Lam, Willcox, and Wolpert]{lam2016bayesian}
Remi Lam, Karen Willcox, and David~H Wolpert.
\newblock Bayesian optimization with a finite budget: An approximate dynamic
  programming approach.
\newblock In \emph{Advances in Neural Information Processing Systems}, pages
  883--891, 2016.

\bibitem[LeCun and Cortes(1998)]{lecun1998mnist}
Y.~LeCun and C.~Cortes.
\newblock The mnist database of handwritten digits, 1998.

\bibitem[Li et~al.(2016)Li, Jamieson, DeSalvo, Rostamizadeh, and
  Talwalkar]{li2016hyperband}
Lisha Li, Kevin Jamieson, Giulia DeSalvo, Afshin Rostamizadeh, and Ameet
  Talwalkar.
\newblock Hyperband: A novel bandit-based approach to hyperparameter
  optimization.
\newblock \emph{arXiv preprint arXiv:1603.06560}, 2016.

\bibitem[MacKay(2003)]{mackay2003information}
David~JC MacKay.
\newblock \emph{Information theory, inference and learning algorithms}.
\newblock Cambridge university press, 2003.

\bibitem[Maclaurin et~al.(2015)Maclaurin, Duvenaud, and
  Adams]{maclaurin2015gradient}
Dougal Maclaurin, David Duvenaud, and Ryan Adams.
\newblock Gradient-based hyperparameter optimization through reversible
  learning.
\newblock In \emph{International Conference on Machine Learning}, pages
  2113--2122, 2015.

\bibitem[Nishihara et~al.(2016)Nishihara, Lopez-Paz, and
  Bottou]{nishihara2016no}
Robert Nishihara, David Lopez-Paz, and L{\'e}on Bottou.
\newblock No regret bound for extreme bandits.
\newblock In \emph{AISTATS}, pages 259--267, 2016.

\bibitem[Ryzhov(2016)]{ryzhov2016convergence}
Ilya~O Ryzhov.
\newblock On the convergence rates of expected improvement methods.
\newblock \emph{Operations Research}, 64\penalty0 (6):\penalty0 1515--1528,
  2016.

\bibitem[Shahriari et~al.(2016)Shahriari, Swersky, Wang, Adams, and
  de~Freitas]{shahriari2016taking}
Bobak Shahriari, Kevin Swersky, Ziyu Wang, Ryan~P Adams, and Nando de~Freitas.
\newblock Taking the human out of the loop: A review of bayesian optimization.
\newblock \emph{Proceedings of the IEEE}, 104\penalty0 (1):\penalty0 148--175,
  2016.

\bibitem[Snoek et~al.(2012)Snoek, Larochelle, and Adams]{snoek2012practical}
Jasper Snoek, Hugo Larochelle, and Ryan~P Adams.
\newblock Practical bayesian optimization of machine learning algorithms.
\newblock In \emph{Advances in neural information processing systems}, pages
  2951--2959, 2012.

\bibitem[Sun et~al.(2008)Sun, Stevens-Navarro, and Wong]{sun2008constrained}
Chi Sun, Enrique Stevens-Navarro, and Vincent~WS Wong.
\newblock A constrained mdp-based vertical handoff decision algorithm for 4g
  wireless networks.
\newblock In \emph{Communications, 2008. ICC'08. IEEE International Conference
  on}, pages 2169--2174. IEEE, 2008.

\bibitem[Swersky et~al.(2014)Swersky, Snoek, and Adams]{swersky2014freeze}
Kevin Swersky, Jasper Snoek, and Ryan~Prescott Adams.
\newblock Freeze-thaw bayesian optimization.
\newblock \emph{arXiv preprint arXiv:1406.3896}, 2014.

\end{thebibliography}
}

\appendix
\section*{Appendices}
%!TEX root = main.tex
\section{Notation Table}
A complete list of notations is summarized in Table.~\ref{table:notation}.
\begin{landscape}
%\section{Notation and Preliminaries}
\bgroup
\everymath{\displaystyle}

\begin{table}[htbp]
\hspace*{-2cm} 
%\centering
\caption{Notations} \label{table:notation}
%\begin{adjustbox}{width=\textwidth}
\centering
\begin{tabular}{c|c|l l}
\hline
section & notation & \multicolumn{2}{c}{ in hyperparameter tuning} \\
\hline
\multirow{6}{*}{\shortstack[l]{problem \\statement}} 
& $[K]$&  set of configurations/arms & \\
& $ B$ & \multicolumn{2}{l}{total budget, sum of epochs from all configurations}    \\
 & $ \vct{b} = (b^1, \ldots, b^k)$ & $b^k$ epochs ran on configuration $k$ &  $\vct{b}\T {\mathbf{1}}_K = B$\\ 
 & $ r $ & remaining budget at step $n$ &  $r = B - n$\\
 & $ y^k(t)$ & loss of configuration $k$ at $t$-th epoch  &  \\ 
 & $\nu^k_{b^k} $ & best/minimum loss of $k$ after $b^k$ epochs & $\nu^k_{b^k}  = \min_{1\leq t \leq b^k} y^k(t)$ \\  
  & $\ell_B $ & final loss & $\ell_B = \min_k \nu^k_{b^k}$ \\  
  & $c$ & optimal configuration & $c = \argmin_k \nu_B^k$ \\
   & $\ell^*_B $ & optimal final loss & $\ell^*_B = \nu_B^c = \min_k \nu_B^k$ \\ 
 \hline
 \multirow{3}{*}{\shortstack[l]{sequential  \\ decision \\ making}} 
 & $ a_n$ & action/configuration selected at the $n$-th step & $a_n \in [K] $  \\
& $ z_n$ & observed loss at the $n$-th step & $z_n = y^{a_n}(t)$ for some $t$ \\ 
& $\xi_n$ & trajectory of configurations and losses & $\xi_n = (a_1, z_1, \ldots, a_n, z_n)$ \\
& $\ell^\pi_B $ &  loss of policy $\pi$ & $\ell^\pi_B = \min_{1\leq n \leq B} z_n$ \\  
\hline
 \multirow{7}{*}{approach} &
$\mS_n$ & belief model at step $n$ & $\mS_n = g(\xi_n)$ a function of the past trajectory \\
& $\nu^k_r $ &  best (future) loss of $k$ with $r$ more epochs & $\nu^k_r =  \min_{1\leq t \leq r} y^k(t_0 + t) $\footnote{Assume $k$ starts from some initial epoch $t_0$.}, r.v. in the belief \\ 
& $\mu^k_{r}$ & expected best loss of $k$ with $r$ more epochs & $\mu^k_r = \expect{}{\nu_r^k }$\\ 
& $\widehat{c}$ & predicted top configuration & $\widehat{c} = \argmin_k \mu_r^k$ \\ 
& $\mu^{\text{1st}}_r$ & 
predicted top configuration loss & $\mu^{\text{1st}}_r =  \min_k \mu_r^k = \mu^{\widehat{c}}_r$ \\ 
& $\mu^{\text{2nd}}_r$ & predicted runner-up configuration loss & $\mu^{\text{2nd}}_r =  \min_{k\neq \widehat{c}} \mu_r^k$\\
& $\tau^\star$ & epochs short from convergence on arm $\widehat{c}$ &  $\tau^\star = \argmin_{1\leq t \leq r} \expect{}{y^{\widehat{c}}(t_0+t)}$ \\ \hline
 \multirow{2}{*}{Others} & $\zeta_n$ & minimum loss from the past & $\zeta_n = \min_{1\leq s \leq n-1} z_s$ \\ \hline
\end{tabular}
%\end{adjustbox}
\end{table}
\egroup
\end{landscape}

\section{Visualization of the action-value function}
\label{ap:visual}
Recall the action value function in Eq.~\ref{eq:Qsub} and~\ref{eq:Qopt},
\begin{align*}
\mQ_r[a] & = \ProbOpr{E}\left[\min \{ \nu^a_{r},  \mu_{r}^{\text{1st}}\} \right]
= \mu_{r}^{\text{1st}} - \ProbOpr{E}\left[\left (\mu_{r}^{\text{1st}} - \nu^a_{r} \right)^+ \right] & a\neq  \widehat{c}.\\
\mQ_r[\widehat{c}] & = \ProbOpr{E}\left[\min \{ \nu^{\widehat{c}}_{r}, \mu_{r}^{\mathrm{2nd}} \}\right]
=  \mu_{r}^{\mathrm{1st}} -  \ProbOpr{E}\left[\left (
\nu^{\widehat{c}}_{r} -  \mu_{r}^{\mathrm{2nd}}
\right)^+ \right], &  a = \widehat{c}.
\end{align*}

\begin {figure}[hbpt]
\centering
\includegraphics[width=0.5 \textwidth]{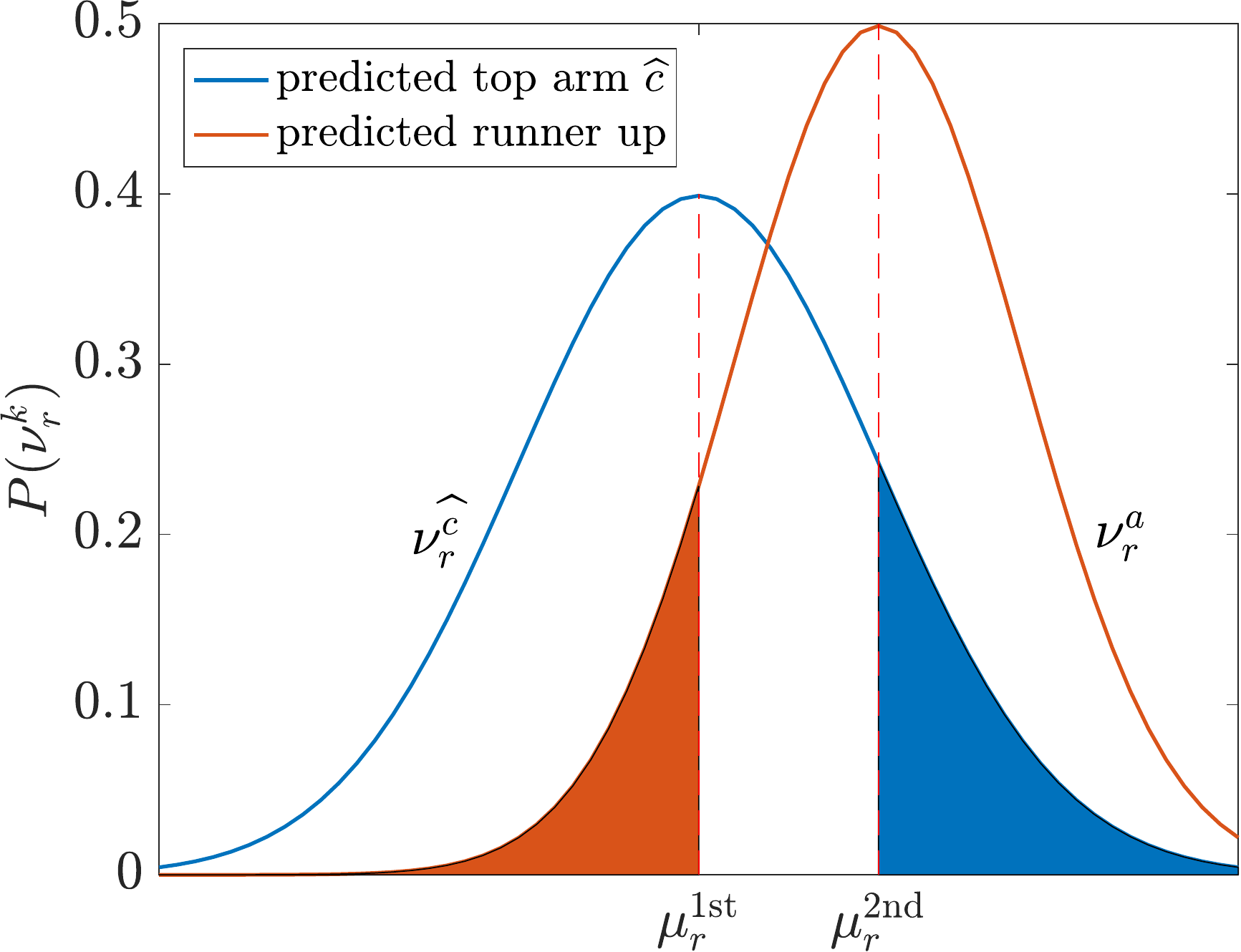}
\caption{Visualization of $\mQ[a]$.} \label{fig:viz_acq}
\end{figure}

If we assume Gaussian distribution for the random variables $\nu_r^k$, $\ProbOpr{E}\left[\left (
\nu^{\widehat{c}}_{r} -  \mu_{r}^{\mathrm{2nd}}
\right)^+ \right]$ and $ \ProbOpr{E}\left[\left (\mu_{r}^{\text{1st}} - \nu^a_{r} \right)^+ \right] $ are the colored area in blue and red respectively in the figure.
%%%%%%%%%%%%%%%%%%%%%%%

\section{Bayes Optimal Solution}
\label{app:dp}
\bgroup
\everymath{\displaystyle}
The Bayes optimal solution optimizes  the following objective,
\begin{align}
\min_{\pi}\expect{0}{\ell^\pi_B } = \min_{\pi}\expect{}{\ell^\pi_B \given \mS_{0}} = \min_{\pi} \expect{}{\min_{1\leq n \leq B} {z_n} \given[\Big] \mS_{0}} \label{eq:sobjBayes},
\end{align}
where $\mS_0$ is the prior of training curves, and we use $\expect{0}{\cdot}$ for abbreviation. To write out the Bellman equation for Eq.~\ref{eq:sobjBayes}. We have to be careful when identifying the optimal sub-structure, since we cannot trivially swap the expectation with the minimum. 

First rewrite Eq.~\ref{eq:sobjBayes} as an accumulated sum of  intermediate improvement, by applying the trick $(c-x)^+ = c - \min\{c, x\}$, 
\begin{align*}
1- \min\{z_1, \ldots, z_{B}\} & = \zeta_1 -  z_1 + \sum_{n=2}^{B}  \min\{z_1, \ldots, z_{n-1}\}  - \min\{z_1, \ldots, z_n\} \\
& = \sum_{n=1}^{B}  \zeta_n - \min\{\zeta_n, z_n\}  = \sum_{n=1}^{B}  (\zeta_n - z_n)^+
\end{align*}
where we define $\zeta_1 = 1$, and $\zeta_n = \min_{1\leq s \leq n-1}z_s$, the best loss in the history prior to step $n$. Then we have the following recursion,
{
\medmuskip=0mu
\thinmuskip=0mu
\thickmuskip=-1mu
%\nulldelimiterspace=0pt
%\scriptspace=0pt
%\vspace*{-\baselineskip}
\begin{align}
\max_{\pi}  \bm{V}_{1} & =  \max_{\pi} \ProbOpr{E}_0 \left[ 1 - \ell_B^\pi \right] 
=\max_{\pi} \ProbOpr{E}_0 \Big[ 1 - \min_{1\leq n \leq B} z_n \Big]. \notag \\
\underbrace{
\ProbOpr{E}_{n-1} \left[ \zeta_n - {\color{blue}\min\{\zeta_n,  \min_{n\leq m \leq B} z_m\}} \right] \vphantom{g}
}_{\bm{V}_n} & = 
\ProbOpr{E}_{n-1} \Big[(\zeta_n - z_n)^+ + 
 \underbrace{ 
\ProbOpr{E}_{n | n-1}  \big[ \zeta_{n+1} -{\color{blue} \min\{\zeta_{n+1},  \min_{n+1\leq m \leq B} z_m\}}\big] 
\vphantom{g} 
}_{\bm{V}_{n+1}} \Big]. \label{eq:Bellman}
\end{align}
}The Bellman equation between the tail value $\bm{V}_n$ and $\bm{V}_{n+1}$ is given in Eq.~\ref{eq:Bellman}.
 $\zeta_n$ is the ``state'' of the system, as it summarizes the past information (best loss), and determines the future reward.
We use $\expect{n-1}{\cdot}$ and $\expect{n|n-1}{\cdot}$ to distinguish different beliefs between steps. 

The computational challenges to solve this DP are two-fold: firstly the reachable states grow exponentially with the remaining horizon, due to the update $\expect{n|n-1}{\cdot}$; 
secondly we do not have the optimal policy when computing the tail value $\bm{V}_{n+1}$.

% Applying the trick  $(c-x)^+ = c - \min\{c, x\}$, 

\paragraph{Approximate Dynamic Programming (Rollout)}
\eat{There is abundant literature of efficient approximate solutions to the Bayes optimality problem (Eq.~\ref{eq:sobjBayes}) in dynamic programming (DP)~\citep{bertsekas1995dynamic}.
In what follows, we describe a rollout approach~\cite{lam2016bayesian} from the Bayesian optimization community, the most applicable to our task.    }

%In this DP, $\zeta_n$ is like the ``state'' of the system, as it summarizes the past information (best loss), and determines the future reward through 
{
%\thickmuskip=-1mu
%\begin{align}
%\min\{\zeta_n,  \min_{n\leq m \leq B} z_m\} =  \zeta_n - \underbrace{\Big( \zeta_n -  \min_{n\leq m \leq B} z_m\Big)^+ \vphantom{g}}_{\text{improvement}}. \label{eq:reward}
%\end{align} 
}
%Note that the particular form of how the past/state $\zeta_n$ influences the reward is non-trivial (Eq.~\ref{eq:reward}). Thus it motivates the \EI heuristics as the rollout policy in the approximation.

The rollout solution proposed in~\cite{lam2016bayesian} is as follows. At step $n$, we expand to rolling horizon $\widetilde{N} = \min\{n+h, B\}$.
{
%\medmuskip=-.5mu
%\thinmuskip=-.5mu
%\thickmuskip=-1mu
%\nulldelimiterspace=0pt
%\scriptspace=0pt
\begin{align}
H_{\widetilde{N}}(\widetilde{\mS}_{\widetilde{N}})  & =\ProbOpr{E} \left[(\zeta_{\widetilde{N}} - z_{\widetilde{N}}^a)^+ \given[\Big] \widetilde{\mS}_{\widetilde{N}} \right] 
= \mathrm{EI} \left[ z_{\widetilde{N}}^a, \zeta_{\widetilde{N}} \given[\Big] \widetilde{\mS}_{\widetilde{N}} \right] \label{eq:rolloutT}  \\
H_k(\widetilde{\mS}_{k})  & = \ProbOpr{E} \left[(\zeta_{k} - z_{k}^{a})^+ 
+ H_{k+1}(\widetilde{\mS}_{k+1})   \given[\Big] \widetilde{\mS}_{k} \right] & \text{ for } k=n+1, \ldots, \widetilde{N}-1.  \notag \\
& \approx  \sum_{q=1} ^{N_q} \alpha^{(q)}
 \left[\left(\zeta_{k} - (z_{k}^{a})^{(q)} \right)^+ 
+ H_{k+1}(\widetilde{\mS}^{(q)}_{k+1})   \right] 
\label{eq:rollout}
\end{align}
}where in Eq.~\ref{eq:rolloutT} the expectation can be computed exactly, and Eq.~\ref{eq:rollout}  is approximated by Gauss-Hermite quadrature. 
$\alpha^{(q)}$ are the  quadrature weights. for $\mathrm{EI}$ is the expected improvement heuristics used in BO community.

\section{Freeze-thaw GP}\label{app:freezethaw}

In this section, we provide details of how to compute the posterior distribution of future predictions in Freeze-Thaw Gaussian process~\cite{swersky2014freeze}. 
We will also write out the log likelihood function, which is used to sample the hyper-parameters for GP in real world experiments.

We use $(x_k,t)$ to index the hyper-parameters $x_k$ and epochs $t$ respectively, and the loss of configuration $k$ from the $t$-th epoch is $y^k(t)$.
The joint distribution of losses from all configurations and epochs, \[\vy = [\vy^1, \ldots, \vy^K] = [y^{1}(1), y^{1}(2), \ldots, y^{1}(n_1),\ldots, y^{K}(1), \ldots, y^{K}(n_K)]\T,\] (arm $k$ has $n_k$ epochs/losses), is given by
\footnote{Assume $N = \sum_{k} n_k$. The dimensionality of variables are: 
{
$\vy \in \R^N, \mK_x \in \R_+^{K\times K}, \mO \in \{0, 1\}^{N\times K}, \ \text{and} \ \mK_{\vt} \in \R_+^{N\times N}$.
}
}
\begin{align}
\Pr (\vy | (\vct{k},\vt))  = \int [\Pi_{k=1}^K \mathcal{N}(\vy_k; f_k \ones_n, \mK_{t_k})] \mathcal{N}(\vct{f}; \vct{m}, \mK_{x}) \mathrm{d}\vct{f}
= \mathcal{N}
\left(
\vy ;
\vct{0} ,
\mK_{\vct{t}} + \mO \mK_x \mO\T
\right).
\label{eq:GP}
\end{align}
$\mO = \mathrm{block} \diag(\bm{1}_{n_1},  \ldots, \bm{1}_{n_K})$ is a block diagonal matrix, where each block is a vector of ones of length $n_k$. 
Kernel $\mK_x$ models the correlation of asymptote losses across \emph{different configurations}, \ie final losses of similar hyper-parameters should be close, as in most conventional hyper-parameter tuning literature. Additionally, kernel
$\mK_{\vct{t}}$ characterizes the correlation of losses from \emph{different epochs} on the same training curve. 
$\mK_{\vct{t}} = \mathrm{block} \diag(\mK_{t_1}, \ldots, \mK_{t_K} )$, the $k$-th block computes the covariance of the losses from arm $k$.
Each entry in $\mK_{t_k} $ is computed via a specific Freeze-Thaw kernel  given by 
\begin{equation*}
k(t, t') = \frac{\beta^\alpha}{(t+t'+\beta)^\alpha}.
\end{equation*} 
It is derived from an infinite mixture of exponentially decaying basis functions, to capture the decay of losses versus time.  Note that conditioned on asymptotes $\vct{f}$, each training curve is drawn independently.

The distribution of the $N$ training curves $\{\vy_n\}_{n=1}^N$ is given by
\begin{equation*}
\begin{aligned}
& \Pr(\{\vy_n\}_{n=1}^N | \{\vx_n\}_{n=1}^N)  \\
 &= \int [\Pi_{n=1}^N \mathcal{N}(\vy_n; f_n\ones_n, \mK_{tn})] \mathcal{N}(\vct{f}; \vct{m}, \mK_{x}) \mathrm{d}\vct{f}
\end{aligned}
\end{equation*}
where we assume each training
curve $\vy_n$ is drawn independently from a Gaussian process with kernel $\mK_{tn}$  conditioned on the prior mean $\vf_n$, which is itself drawn
from a global Gaussian process with kernel $\mK_x$ and mean $m$. $\mK_x$ can be any common covariance function which models the correlation of performances of different hyper-parameters.

We summarize and introduce notations as follows,
\begin{align*}
\mO & = \mathrm{block} \diag(\bm{1}_{n_1}, \bm{1}_{n_2}, \ldots, \bm{1}_{n_N}) \\
\mK_t & =  \mathrm{block} \diag(\mK_{t1}, \ldots, \mK_{t_N}) \\
\vy & = (\vy_1, \ldots, \vy_N)\T \\
\Lambda & = \mO \T \mK_t^{-1}\mO \\
\gamma & = \mO \T \mK_t^{-1} (\vy - \mO \vct{m}) =   \mO \T \mK_t^{-1} \vy - \Lambda \vct{m} \\
\mat{C} & = (\mK_x^{-1} + \Lambda)^{-1} \\
\mu & = \vct{m} + \mat{C} \gamma
%\eta & = \mO\T \mK_t^{-1} \vy = \gamma + \Lambda \vm \\
%\vz & = \Lambda^{-1} \eta = (\mO \T \mK_t^{-1} \mO)^{-1}\mO\T \mK_t^{-1} \vy = \Lambda^{-1} \gamma + \vm \\
%\mK_{z} &= \mK_{x} + \Lambda^{-1}
\end{align*}

\eat{
where $\mO$ is a block-diagonal matrix of $\ones_{n_n}$ vectors of length $n_n$, and $\mK_t$ is a block-diagonal matrix with $\mK_{tn}$ on the diagonal block.}

The log likelihood of the Gaussian process is given by
\begin{equation*}
\begin{aligned}
& \log P(\vy | \{\vx_k\}_{k=1}^K) \\
 & = -\frac{1}{2}  (\vy - \mO \vct{m})\T \mK_t^{-1} (\vy - \mO \vct{m}) \\
 & + \frac{1}{2}\gamma\T (\mK_x^{-1} + \Lambda)^{-1} \gamma
- \frac{1}{2} (\log |\mK_x^{-1}+ \Lambda|) \\
&+ \log (|\mK_x|)  + \log (|\mK_t|) + \text{const}.
\end{aligned}
\end{equation*}
The posterior distribution of the predicted mean of the seen and unseen curves are given by
\begin{equation*}
\begin{aligned}
%\Pr(\vf | \vy, \{\vx_n\}_{n=1}^N ) = \mathcal{N} (\vf; & \vm + (\mK_x^{-1} + \Lambda)^{-1} \gamma, \\
%& (\mK_x^{-1} + \Lambda)^{-1} ) \\
\Pr(\vct{f} | \vy, \{\vx_k\}_{k=1}^K )& = \mathcal{N} (\vct{f};  \vct{m} + \mat{C} \gamma, \mat{C}), \\
\Pr(f _*| \vy, \{\vx_k\}_{k=1}^K, \vx_* ) & =  \mathcal{N} (f_*;  m + \mK_{x*}\T\mK_x^{-1}\mat{C }\gamma, \\
& \mK_{x**} - \mK_{x*}\T(\mK_x^{-1} + \Lambda^{-1})^{-1}\mK_{x*})
\end{aligned}
\end{equation*}
respectively.
The posterior distribution for a new observation on a seen training curve is given by
\begin{equation*}
\begin{aligned}
\Pr(y_{n*}| \vy, \{\vx_k\}_{k=1}^K, t_*) & =  \mathcal{N} (y_{n*};  \mK_{tn*}\T\mK_{tn}^{-1}\vy_n + \Omega \mu_n,\\
& \mK_{tn**} - \mK_{tn*}\T\mK_{tn}^{-1}\mK_{tn*} + \Omega\mat{C}_{nn}\Omega\T)
\end{aligned}
\end{equation*}
where $\Omega = \ones_{*} - \mK_{tn*}\T\mK_{tn}^{-1}\mK_{tn*}$.
And the posterior distribution for a new observation on an unseen curve is given by
\begin{equation*}
\begin{aligned}
& \Pr(y_{n*}| \vy, \{\vx_k\}_{k=1}^K, \vx_*, t_*)  =  \mathcal{N} (y_{n*};  m + \mK_{x*}\T\mK_x^{-1}\mat{C} \gamma, \\
& \mK_{t**} +  \mK_{x**} - \mK_{x*}\T(\mK_x^{-1} + \Lambda^{-1})^{-1}\mK_{x*}).
\end{aligned}
\end{equation*}

\section{Experimental Details}\label{app:exp} 
In this section, we provide more details of the experiment.

\paragraph{Data}
We create  synthetic data for budgeted optimization, and hyperparameter tuning tasks on well-established real-world datasets, \cifar~\citep{krizhevsky2009learning} and MNIST~\citep{lecun1998mnist}. 

\begin{enumerate}
\item \textbf{Synthetic loss functions.} 
We generate 100 synthetic sets of time-varying loss functions, which resemble the learning curves  structure over configurations in hyperparameter tuning. Each of the synthetic set consists of $K=84$ configurations with maximum length 288 training epochs. 

The configurations' converged performances are drawn from a zero-mean magnitude 1 Gaussian process with squared exponential kernel of length-scale $0.8$. For each configuration, the loss decay curve is sampled from the Freeze-Thaw kernel with $\alpha = 1.5, \beta = 5$, and magnitude $10$. The budget unit is 6 epochs.
\end{enumerate}
For real-world data experiments, we create 4 hyperparameter tuning tasks using \cifar and MNIST datasets. We tune hyperparameters of convolutional neural networks on \cifar dataset~\citep{krizhevsky2009learning}. We split the data into $40k, 10k$ for training and heldout sets, and report the error rate on the heldout set. We create two hyperparmater tuning tasks on \cifar.
All the CNN models are trained on Tesla K20m GPU using Tensorflow. 
\begin{enumerate}[resume]
\item \textbf{ResNet on \cifar.} We randomly sample 96 configurations of 5 hyperparameters of ResNet~\citep{he2016deep} model as follows, optimizers from $\{$stochastic gradient descent, momentum gradient descent, adagrad$\}$, batch size in $\{64, 128\}$, learning rate in $\{0.01, 0.05, 0.1, 0.5, 1.0\}$, momentum in $\{0.5, 0.9\}$, and different exponential learning rate decay rates. The budget unit is approximately 16 minute of GPU training time for this task.

\item \textbf{AlexNet and ResNet on \cifar.} We tune the architectural selection between ResNet~\citep{he2016deep} and AlexNet~\citep{krizhevsky2012imagenet} as well as other hyperparameters in this task. We have 49 candidate configurations evenly split between the two architectures, with 24 ResNet and 25 AlexNet. The range of other 5 hyperparameters is the same as the ones described in (ii). The budget unit is approximately 14 minute of GPU training time on this task.
\end{enumerate}
For hyperparameter tuning tasks on MNIST, we use openly available dataset from \textsf{LC-dataset}\footnote{\url{http://ml.informatik.uni-freiburg.de/people/klein/index.html}} ~\citep{klein2017learning}, where training curves of different hyperparameters are provided. We choose the unsupervised learning task of learning image distribution using variational autoencoder (VAE), and classification task by fully connected neural net. Please see~\citep{klein2017learning} for more details on the generation of the learning curves.
\begin{enumerate}[resume]
\item \textbf{VAE on MNIST.}
We subsample 49 configurations of 4 different hyperparameters for training VAE  on MNIST.~\citep{klein2017learning}  The loss observations are lower bound of the heldout log-likelihood. The budget unit is 10 epochs. \eat{We calibrate the value of log-likelihood globally by subtracting a constant (an estimate of the lower bound of all values) and divided by the maximum of all values, for computational convenience.\footnote{Note that we didn't calibrate each curve individually which require the information not available before training. The estimate of the lower bound of all values needs not be accurate.}}

\item \textbf{FCNet on MNIST.}
We subsample 50 configurations of 10 different hyperparameters for classification using two layer fully connected network on MNIST. The loss observations are error rates on heldout set. The budget unit is 10 epochs.
\end{enumerate}

\paragraph{Implementation Details}
%To have a parameter-free tuning algorithm, we specify $\beta = 0.5$ in score2 (Eq.~\ref{eq:acq2}) in our experiments, following the practice in~\citep{qin2017improving}. 
On synthetic task for both our method and GP-EI, we use the same GP hyperparameters as the ones in data generation. 
For real-world tasks, we use Freeze-Thaw GP in our algorithm. We assume independence between different configurations for computational efficiency. The hyperparameters of the GP are sampled using slice sampling~\citep{mackay2003information} in a fully-Bayesian treatment. We find it in general not sensitive to the sampling parameter. We didn't tune it and set it to be step size 0.5, burn in samples 10, and max attempts 10.

We implemented Hyperband with $\eta=3$ as recommended by the author\footnote{We have also tried other $\eta$ values, but didn't find significant difference in performance.}. 
For SMAC and Fabolas, we adapt the implementations from \url{https://github.com/automl/}.
For the Rollout~\cite{lam2016bayesian} method, we use rolling horizon $h=3$ and use Gauss-Hermite quadrature to approximate the imaginary belief updates.

% Fabolas
We adapt Fabolas~\citep{klein2017fast} in the following way. The original Fabolas adaptively selects the training subset size, as well as the configurations to by an acquisition function, which trades off information gain with computation cost. 
We interpret the intermediate results from the curve as the subset training result, and input to Fabolas, to decide which configurations to run next and for how many epochs\footnote{In another word, we use epoch, instead of subset size, as the additional input to the black-box optimization problem. The cost of the surrogate task grows linearly w.r.t. the epoch, and can be modeled by the kernel for the computation cost in their work.}.

\subsection*{Other results and analysis} 
\paragraph{On budget exhaustion heuristics}

To  understand the impact  of budget exhaustion heuristics in our algorithm, we compare our methods (BHPT) with $Q$ and $Q-\varepsilon$ only (without the $\cst{else}$ statement in Alg.~\ref{alg}), in light blue and red correspondingly in Fig.~\ref{rebuttal_syn}(a).
$Q$-alone and $Q-\varepsilon$-alone select configurations by the action-value functions (Eqn.~\ref{eq:acq}, Eq.~\ref{eq:policy2}) alone, without the heuristics of committing to improve the top configuration when the budget runs out.  The performance deteriorates, especially for $Q$-alone. 
It has less impact on is $Q-\varepsilon$-alone due to that $Q-\varepsilon$ has already had sufficient eploitation ($\varepsilon$-greedy, where $\varepsilon = 0.5$) in its design.

%Meanwhile, the small differences between two red curves in Fig.~\ref{syn}(b) is consistent with the observations in resource allocation in Fig.~\ref{alloc}(a).

 \paragraph{On full Bayesian treatment of GP hyper-parameters}
In Fig.~\ref{rebuttal_syn}(b), we demonstrate the effect of doing full Bayesian sampling on the hyper-parameters of the Freeze-Thaw GP versus a fixed GP hyper-parameter on the synthetic data set.
\begin {figure}[hbtp]%[!hbtp]
\centering
%\begin{adjustbox}{width=0.5\textwidth}
%\hspace{-.6in}
\begin{tabular}{cc}
 \includegraphics[width=0.45\textwidth]{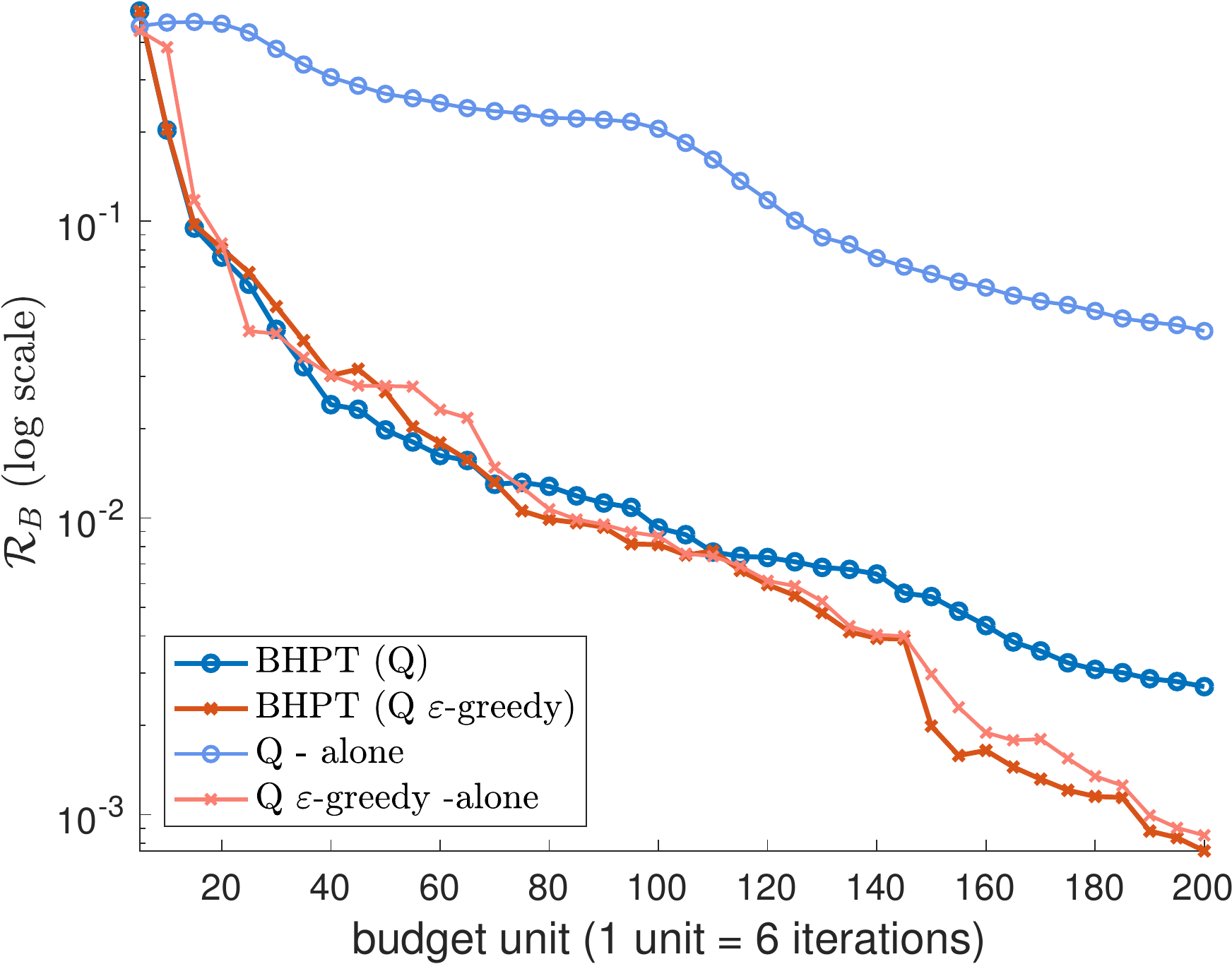}
 & 
 \includegraphics[width=0.45\textwidth]{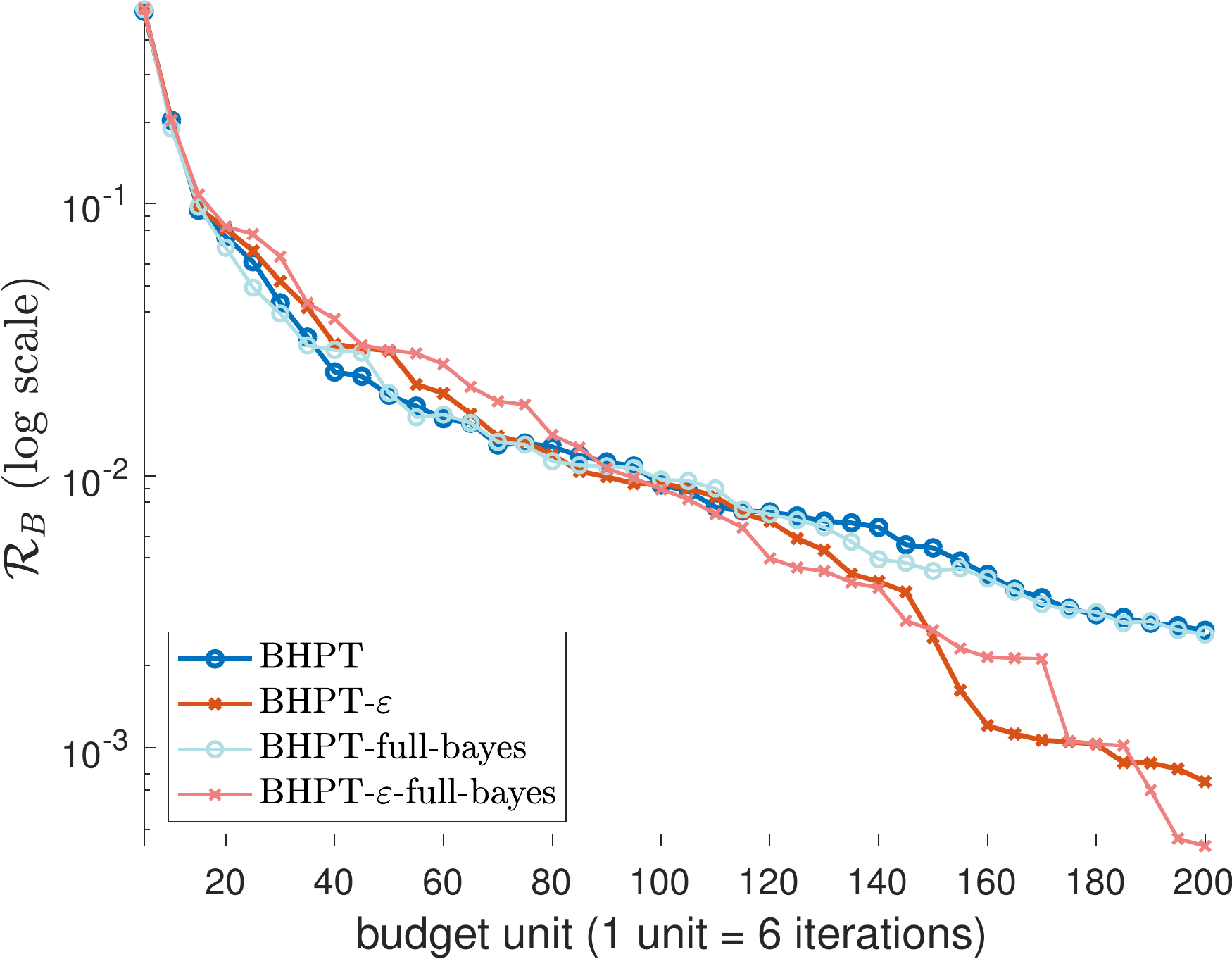}
 \\  \hspace{-.3in}
 {(a)  BHPT without budget exhaustion heuristic} 
 &\hspace{-.1in}
  {(b) full Bayesian treatment of the GP hyper-parameters } \\
  {$Q$ - alone does less exploitation than BHPT, }
  & {Full Bayesian treatment is similar to GP }\\
  {thus results in worse output performance.} 
  &
  {hyper-parameters  set to the ground-truth values.}
\end{tabular}
%\end{adjustbox}
\caption{More Analysis on Budgeted Optimization on 100 Synthetic Sets}
\label{rebuttal_syn}
\end{figure}

%\paragraph{Baseline methods}
%
%The baseline methods we compare to include the mainstream hyperparameter tuning methods, SMAC~\citep{hutter2011sequential} in Bayesian optimization and Hyperband~\citep{li2016hyperband}, and a state-of-the-art one Fabolas~\citep{klein2017fast}.}
%As we are the first to propose the budgeted setting for hyperparameter tuning, we also implemented naive heuristics as baselines to compare to our method, and justify the design of our algorithm. 
%(d) \textbf{myopic}. One straightforward solution is to apply the conventional Expected Improvement (EI) acquisition function in BO to every unit resource.
%
%\begin{align}
%i_n = \argmax_{a\in[K]}\expect{n}{(\tau_n - z_n(a))^+}.
%\end{align}
%
%Compared to our action-value functions, Eqn.~\ref{eq:myopic} does not use forecast long term predictions of configurations w.r.t. $\mR_n$, but focuses on the next step loss near-sightedly instead. 
%
%(e) \textbf{score1-if, score2-if}.
%Another baseline is to select configurations by the action-value functions proposed in Sect.~\ref{sect:acq} and stay in the \texttt{if} branch in Alg.~\ref{alg:framework} throughout tuning. This baseline does not address the challenge of resource competition between exploration and future improving. \eat{, and ignorantly focuses on the exploration and improving tradeoff alone.} We refer to these baselines as score1-if and score2-if.
%
%

\end{document}